\documentclass[sigconf]{acmart} 

\AtBeginDocument{%
  \providecommand\BibTeX{{%
    \normalfont B\kern-0.5em{\scshape i\kern-0.25em b}\kern-0.8em\TeX}}}

\usepackage{latexsym}
\usepackage{amsmath}
\usepackage{url}
\usepackage{amsmath,nccmath}
\usepackage{algorithm}
\usepackage{algorithmic}
\usepackage{xcolor}
\usepackage{graphicx}
\usepackage{subfigure}
\usepackage{wrapfig}
 \usepackage{relsize}
\usepackage{balance}
\usepackage{bbm}

\theoremstyle{definition}

\usepackage{enumitem}

\usepackage{array}
\newcolumntype{P}[1]{>{\centering\arraybackslash}p{#1}}

\usepackage{multirow}

\copyrightyear{2023} 
\acmYear{2023} 
\setcopyright{acmcopyright}\acmConference[WSDM '23]{Proceedings of the
Sixteenth ACM International Conference on Web Search and Data
Mining}{February 27-March 3, 2023}{Singapore, Singapore}
\acmBooktitle{Proceedings of the Sixteenth ACM International Conference on
Web Search and Data Mining (WSDM '23), February 27-March 3, 2023,
Singapore, Singapore}
\acmPrice{15.00}
\acmDOI{10.1145/3539597.3570421}
\acmISBN{978-1-4503-9407-9/23/02}

\title{Towards Faithful and Consistent Explanations for Graph Neural Networks
}

 \author{Tianxiang Zhao}
\affiliation{%
  \institution{College of Information Sciences and Technology, The Pennsylvania State University}
  \city{State College}
  \country{USA}}
\email{tkz5084@psu.edu}

 \author{Dongsheng Luo}
\affiliation{%
  \institution{Knight Foundation School of Computing and Information Sciences, Florida International University}
  \city{Miami}
  \country{USA}}
\email{dluo@fiu.edu}

 \author{Xiang Zhang}
\affiliation{%
  \institution{College of Information Sciences and Technology, The Pennsylvania State University}
  \city{State College}
  \country{USA}}
\email{xzz89@psu.edu}

 \author{Suhang Wang}
\affiliation{%
  \institution{College of Information Sciences and Technology, The Pennsylvania State University}
  \city{State College}
  \country{USA}}
\email{szw494@psu.edu}

\renewcommand{\shortauthors}{Tianxiang Zhao, Dongsheng Luo, Xiang Zhang, \& Suhang Wang}

\begin{document}
\begin{abstract}
Uncovering rationales behind predictions of graph neural networks (GNNs) has received increasing attention over recent years. Instance-level GNN explanation aims to discover critical input elements, like nodes or edges, that the target GNN relies upon for making predictions. 
Though various algorithms are proposed, most of them formalize this task by searching the minimal subgraph which can preserve original predictions. However, an inductive bias is deep-rooted in this framework: several subgraphs can result in the same or similar outputs as the original graphs. Consequently, they have the danger of providing spurious explanations and fail to provide consistent explanations. Applying them to explain weakly-performed GNNs would further amplify these issues. To address this problem, we theoretically examine the predictions of GNNs from the causality perspective. Two typical reasons of spurious explanations are identified: confounding effect of latent variables like distribution shift, and causal factors distinct from the original input. Observing that both confounding effects and diverse causal rationales are encoded in internal representations, we propose a simple yet effective countermeasure by aligning embeddings. Concretely, concerning potential shifts in the high-dimensional space, we design a distribution-aware alignment algorithm based on anchors. This new objective is easy to compute and can be incorporated into existing techniques with no or little effort. 
Theoretical analysis shows that it is in effect optimizing a more faithful explanation objective in design, which further justifies the proposed approach. 
 
\end{abstract}



\maketitle

\section{Introduction}
Graph-structured data is ubiquitous in the real world, such as social networks~\cite{Fan2019GraphNN,zhao2020semi}, molecular structures~\cite{Mansimov2019MolecularGP,Chereda2019UtilizingMN} and knowledge graphs~\cite{Sorokin2018ModelingSW}. With the growing interest in learning from graphs, graph neural networks (GNNs) are receiving more and more attention over the years. Generally, GNNs adopt message-passing mechanisms, which recursively propagate and fuse messages from neighbor nodes on the graphs. GNNs have achieved state-of-the-art performance for many tasks such as node classification~\cite{kipf2016semi,velivckovic2017graph,hamilton2017inductive}, graph classification~\cite{xu2018powerful}, and link prediction~\cite{zhang2018link}.

Despite the success of GNNs for various domains, as with other neural networks, GNNs lack interpretability. 
Understanding the inner working of GNNs can bring several benefits. First, it enhances practitioners' trust in the GNN model by enriching their understanding of the network characteristics. Second, it increases the models' transparency to enable trusted applications in decision-critical fields sensitive to fairness, privacy and safety challenges, such as healthcare and drug discovery~\cite{rao2021quantitative}. Thus, studying the explainability of GNNs is attracting increasing attention. 

Particularly, we focus on post-hoc instance-level explanations. Given a trained GNN and an input graph, 
this task seeks to discover the substructures that can explain the prediction behavior of the GNN model. Some solutions have been proposed in existing works~\cite{ying2019gnnexplainer,huang2020graphlime,vu2020pgm}. For example, in search of important substructures that predictions rely upon, GNNExplainer learns an importance matrix on nodes and edges via perturbation~\cite{ying2019gnnexplainer}. The identified minimal substructures that preserve original predictions are taken as the explanation. Extending this idea, PGExplainer trains a graph generator to utilize global information in explanation and enable faster inference in the inductive setting~\cite{luo2020parameterized}. SubgraphX constraint explanations as connected subgraphs and conduct Monte Carlo tree search based on Shapley value~\cite{yuan2021explainability}. These methods can be summarized in a label preserving framework, i.e., candidate explanation is formed as a masked version of the original graph and is identified as the minimal discriminative substructure that preserves the predicted label.

\begin{figure}[t!]
  \centering
  \subfigure[Run 1]{
		\includegraphics[width=0.23\textwidth]{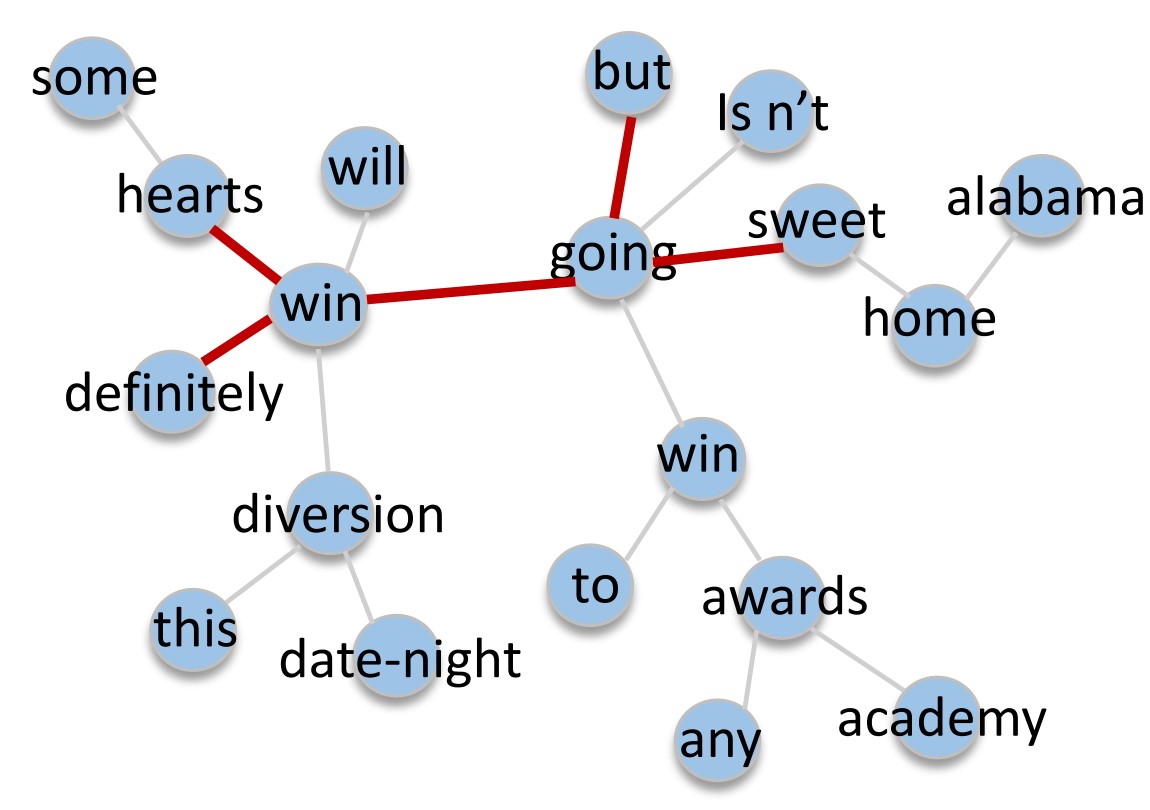}}
  \subfigure[Run 2]{
		\includegraphics[width=0.23\textwidth]{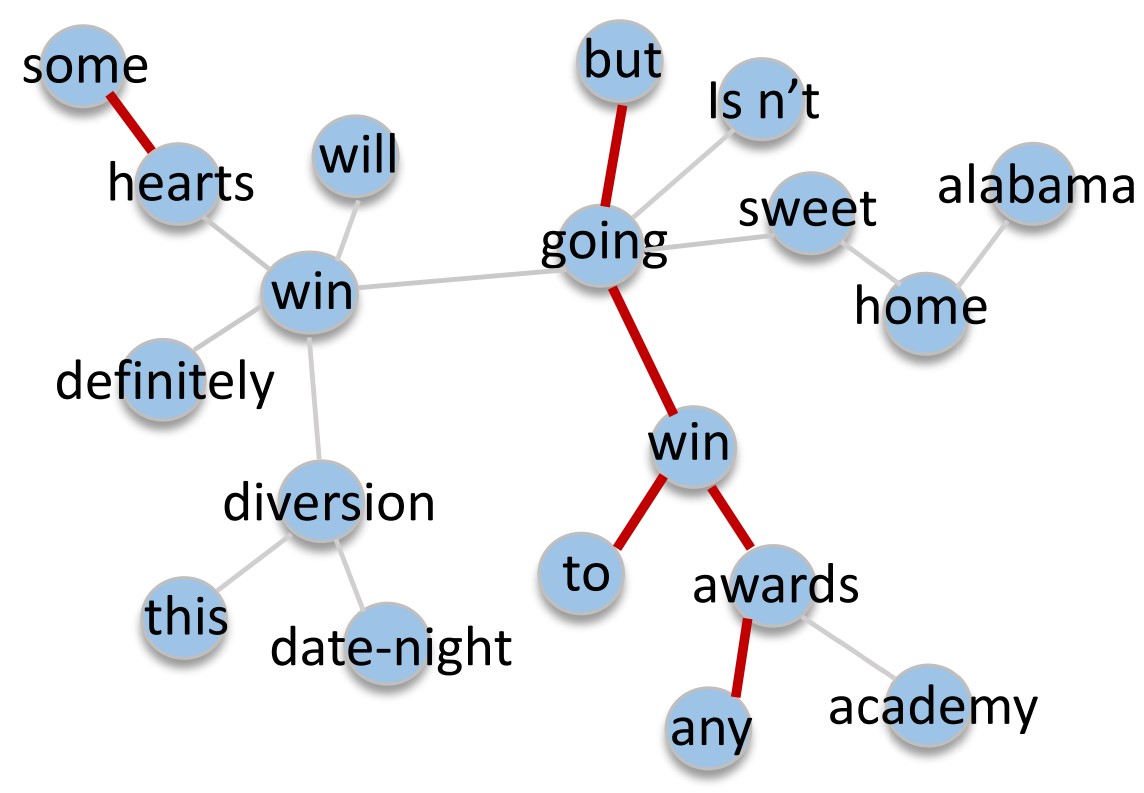}}
    \vskip -1.5em
    \caption{Explanation results achieved by a leading baseline GNNExplainer on the same input graph from Graph-SST5. Red edges formulate  explanation substructures.} \label{fig:case_study}
    \vskip -2.5em
\end{figure}

However, due to the complexity of topology and the combinatory number of candidate substructures, existing label preserving methods are insufficient for a faithful and consistent explanation of GNNs. They are unstable and are prone to give spurious correlations as explanations.
A failure case is shown in Fig.~\ref{fig:case_study}, where a GNN is trained on Graph-SST5~\cite{yuan2020explainability} for sentiment classification. Each node represents a word and each edge denotes syntactic dependency between nodes. Each graph is labeled based on the sentiment of the sentence. In the figure, the sentence ``Sweet home alabama isn't going to win any academy awards, but this date-night diversion will definitely win some hearts'' is labeled \textit{positive}. In the first run, GNNExplainer~\cite{ying2019gnnexplainer} identifies the explanation as ``definitely win some hearts'', and in the second run, it identifies ``win academy awards'' to be the explanation instead. Different explanations obtained by GNNExplainer break the criteria of \textbf{consistency}, i.e., the explanation method should be deterministic and consistent with the same input for different runs~\cite{nauta2022anecdotal}. Consequently, explanations provided by the existing methods may fail to faithfully reflect the decision mechanism of the to-be-explained GNN. 



Inspecting the inference process of target GNNs, we find that the inconsistency problem and spurious explanations can be understood from the causality perspective. Specifically, existing explanation methods may lead to spurious explanations either as a result of different causal factors or due to the confounding effect of distribution shifts (identified subgraphs may be out of distribution). These failure cases originate from a particular inductive bias that predicted labels are sufficiently indicative for extracting critical input components. This underlying assumption is rooted in optimization objectives adopted by existing works~\cite{ying2019gnnexplainer,luo2020parameterized,yuan2021explainability}. However, our analysis demonstrates that the label information is insufficient to filter out spurious explanations, leading to inconsistent and unfaithful explanations. 

Considering the inference of GNNs, both confounding effects and distinct causal relationships can be reflected in the internal representation space. With this observation, we propose a novel objective that encourages alignment of embeddings of raw graph and identified subgraph in internal embedding space to obtain more faithful and consistent GNN explanations. 
Concretely, we propose two strategies, a distribution-aware method based on anchors and a simplified one based on absolute distance. Both variants are flexible to be incorporated into various existing GNN explanation methods. 
Further analysis shows that the proposed method is in fact optimizing a new explanation framework, which is more faithful in design. We summarized our contributions as follows.
\begin{itemize}[leftmargin=*]
    \item We point out the faithfulness and consistency issues in rationales identified by existing GNN explanation models. These issues arise due to the inductive bias in their label-preserving framework, which only uses predictions as the guiding information.
    \item 
    We propose an effective and easy-to-apply countermeasure by aligning intermediate embeddings, which is flexible to be incorporated into various GNN explanation techniques. 
    We further conduct theoretical analysis to understand and validate the proposed framework. 
    \item Extensive experiments on real-world and synthetic datasets show that our framework benefits various GNN explanation models to achieve more faithful and consistent explanations. 
\end{itemize}

\section{Related Work}

\subsection{Graph Neural Networks}
Graph neural networks (GNNs) are developing rapidly in recent years, with the increasing need for learning on relational data structures~\cite{Fan2019GraphNN,dai2022towards,zhao2021graphsmote}. Generally, 
existing GNNs can be categorized into two categories, i.e., spectral-based approaches~\cite{bruna2013spectral,Tang2019ChebNetEA,kipf2016semi} based on graph signal processing theory, and spatial-based approaches~\cite{duvenaud2015convolutional,atwood2016diffusion,xiao2021learning} relying upon neighborhood aggregation. Despite their differences, most GNN variants can be summarized with the message-passing framework, which is composed of pattern extraction and interaction modeling within each layer~\cite{gilmer2017neural}. Specifically, GNNs model messages from node representations. These messages are then propagated with various message-passing mechanisms to refine node representations, which are then utilized for downstream tasks~\cite{hamilton2017inductive,zhang2018link,zhao2021graphsmote}. Explorations are made by disentangling the propagation process~\cite{wang2020disentangled,zhao2022exploring,xiao2022decoupled} or utilizing external prototypes~\cite{lin2022prototypical,xu2022hp}. Despite their success in network representation learning, GNNs are uninterpretable black box models. It is challenging to understand their behaviors even if the adopted message passing mechanism and parameters are given. Besides, unlike traditional deep neural networks where instances are identically and independently distributed, GNNs consider node features and graph topology jointly, making the interpretability problem more challenging to handle. 

\subsection{GNN Interpretation Methods}
Recently, some efforts have been taken to interpret GNN models and provide explanations for their predictions~\cite{wang2020causal}. Existing methods can be generally grouped into two categories based on the granularity: (1) instance-level explanation~\cite{ying2019gnnexplainer}, which provides explanations on the prediction for each instance by identifying important substructures; and (2) model-level explanation~\cite{baldassarre2019explainability,zhang2021protgnn}, which aims to understand global decision rules captured by the target GNN. From the methodology perspective, existing methods can be categorized as (1) self-explainable GNNs~\cite{zhang2021protgnn,dai2021towards}, where the GNN can simultaneously give prediction and explanations on the prediction; and (2) post-hoc explanations~\cite{ying2019gnnexplainer,luo2020parameterized,yuan2021explainability}, which adopt another model or strategy to provide explanations of a target GNN. As post-hoc explanations are model-agnostic, i.e., can be applied for various GNNs, in this work, we focus on post-hoc instance-level explanations~\cite{ying2019gnnexplainer}, i.e., given a trained GNN model, identifying instance-wise critical substructures for each input to explain the prediction. A comprehensive survey can be found in ~\cite{dai2022comprehensive}.

A variety of strategies for post-hoc instance-level explanations have been explored in the literature, including utilizing signals from gradients~\cite{pope2019explainability,baldassarre2019explainability}, perturbed predictions~\cite{ying2019gnnexplainer,luo2020parameterized,yuan2021explainability,shan2021reinforcement}, and decomposition~\cite{baldassarre2019explainability,schnake2020higher}. Among these methods, perturbed prediction-based methods are the most popular. The basic idea is to learn a perturbation mask that filters out non-important connections and identifies dominating substructures preserving the original predictions~\cite{yuan2020explainability}. The identified important substructure is used as an explanation for the prediction. For example, GNNExplainer~\cite{ying2019gnnexplainer} employs two soft mask matrices on node attributes and graph structure, respectively, which are learned end-to-end under the maximizing mutual information (MMI) framework. PGExplainer~\cite{luo2020parameterized} extends it by incorporating a graph generator to utilize global information. It can be applied in the inductive setting and prevent the onerous task of re-learning from scratch for each to-be-explained instance. SubgraphX~\citep{yuan2021explainability} employs Monte Carlo Tree Search (MCTS) to find connected subgraphs that preserve predictions as explanations.

Despite progress in interpreting GNNs, most of these methods discover critical substructures merely upon the change of outputs given perturbed inputs. Due to this underlying inductive bias, existing label-preserving methods are heavily affected by spurious correlations caused by confounding factors in the environment. On the other hand, by aligning intermediate embeddings in GNNs, our method alleviates the effects of spurious correlations on interpreting GNNs, leading to faithful and consistent explanations.

\section{Preliminary}

\subsection{Problem Definition}
We use $\mathcal{G} = \{\mathcal{V},\mathcal{E}; \mathbf{F}, \mathbf{A} \}$ to denote a graph, where $\mathcal{V}=\{v_1,\dots,v_{n}\}$ is a set of $n$ nodes and $\mathcal{E} \in \mathcal{V} \times \mathcal{V}$ is the set of edges. Nodes are accompanied by an attribute matrix $\mathbf{F} \in \mathbb{R}^{n \times d}$, and $\mathbf{F}[i,:] \in \mathbb{R}^{1 \times d}$ is the $d$-dimensional node attributes of node $v_i$. $\mathcal{E}$ is described by an adjacency matrix $\mathbf{A} \in \mathbb{R}^{n \times n}$. ${A}_{ij}=1$ if there is an edge between node $v_i$ and $v_j$; otherwise, ${A}_{ij}=0$. For \textit{graph classification}, each graph $\mathcal{G}_i$ has a label $Y_i \in \mathcal{C}$, and a GNN model $f$ is trained to map $\mathcal{G}$ to its class, i.e., $f: \{\mathbf{F}, \mathbf{A}\} \mapsto \{1, 2, \dots, C \}$. Similarly, for \textit{node classification}, each graph $\mathcal{G}_i$  denotes a $K$-hop subgraph centered at node $v_i$ and a GNN model $f$ is trained to predict the label of $v_i$ based on node representation of $v_i$ learned from $\mathcal{G}_i$. The purpose of explanation is to find a subgraph $\mathcal{G}'$, marked with binary importance mask $\mathbf{M}_{A} \in [0,1]^{n \times n}$ on adjacency matrix and $\mathbf{M}_{F} \in[0,1]^{n \times d}$ on node attributes, respectively, e.g., $\mathcal{G}'=\{\mathbf{A}\odot \mathbf{M}_A; \mathbf{F}\odot \mathbf{M}_F\}$, where $\odot$ denote elementwise multiplication. 
These two masks highlight components of $\mathcal{G}$ that are important for $f$ to predict its label. With the notations, the \textit{post-hoc instance-level} GNN explanation task is defined as:

\vspace{0.5em}
\noindent{}\textit{Given a trained GNN model $f$, for an arbitrary input graph $\mathcal{G}= \{\mathcal{V}, \mathcal{E};  \mathbf{F}, \mathbf{A}\}$, find a subgraph $\mathcal{G}'$ that can explain the prediction of $f$ on $\mathcal{G}$. Obtained explanation $\mathcal{G}'$ is depicted by importance mask $\mathbf{M}_{F}$ on node attributes and importance mask $\mathbf{M}_{A}$ on adjacency matrix.
}

\subsection{ MMI-based Explanation Framework}
Many approaches have been designed for post-hoc instance-level  GNN explanation. Due to the discreteness of edge existence and non-grid graph structures, most works apply a perturbation-based strategy to search for explanations. Typically, they can be summarized as Maximization of Mutual Information (MMI) between predicted label $\hat{Y}$ and explanation $\mathcal{G}'$, i.e.,
\begin{equation}\label{eq:framework}
    \begin{aligned}
    \min_{\mathcal{G}'} & - I(\hat{Y}, \mathcal{G}'), \\
    \quad \text{s.t.} \quad \mathcal{G}' \sim & \mathcal{P}(\mathcal{G}, \mathbf{M}_{A},  \mathbf{M}_{F}),  \quad \mathcal{R}(\mathbf{M}_{F},\mathbf{M}_{A}) \leq c 
    \end{aligned}
\end{equation}
where $I()$ represents mutual information and $\mathcal{P}$ denotes the perturbations on original input with importance masks $\{\mathbf{M}_{F},\mathbf{M}_{A}\}$. For example, let $\{\hat{\mathbf{A}},\hat{\mathbf{F}}\}$ represent the perturbed $\{\mathbf{A}, \mathbf{F}\}$. Then, $\hat{\mathbf{A}}= \mathbf{A} \odot \mathbf{M}_{A}$ and $\hat{\mathbf{F}} = \mathbf{Z} + (\mathbf{F}-\mathbf{Z}) \odot \mathbf{M}_{F}$ in GNNExplainer~\cite{ying2019gnnexplainer}, where $\mathbf{Z}$ is sampled from marginal distribution of node attributes $\mathbf{F}$. $\mathcal{R}$ denotes regularization terms on the explanation, imposing prior knowledge into the searching process, like constraints on budgets or connectivity distributions~\cite{luo2020parameterized}. Mutual information $I(\hat{Y}, \mathcal{G}')$ quantifies consistency between original predictions $\hat{Y}=f(\mathcal{G})$ and prediction of candidate explanation $f(\mathcal{G}')$, which promotes the positiveness of found explanation $\mathcal{G}'$. 
Since mutual information measures the predictive power of $\mathcal{G}'$ on $Y$, this framework essentially tries to find a subgraph that can best predict the original output $\hat{Y}$. During training, a relaxed version~\cite{ying2019gnnexplainer} is often adopted as: 
\begin{equation}\label{eq:surrogate}
    \begin{aligned}
    \min_{\mathcal{G}'} ~& {H}_{C}\big(\hat{Y}, P(\hat{Y}' \mid \mathcal{G}') \big),\\
     \quad \text{s.t.} \quad \mathcal{G}' \sim & \mathcal{P}(\mathcal{G}, \mathbf{M}_{A},  \mathbf{M}_{F}) , \quad \mathcal{R}(\mathbf{M}_{F},\mathbf{M}_{A}) \leq c 
    \end{aligned}
\end{equation}
where $H_C$ denotes cross-entropy. With this same objective, existing methods mainly differ from each other in searching strategies.


Different aspects regarding the quality of explanations can be evaluated~\cite{nauta2022anecdotal}. Among them, two most important criteria are \textbf{faithfulness} and \textbf{consistency}. Faithfulness measures the descriptive accuracy of explanations, indicating how truthful they are compared to behaviors of the target model. Consistency considers explanation invariance, which checks that identical input should have identical explanations. However, as shown in Figure~\ref{fig:case_study}, the existing MMI-based framework is sub-optimal in terms of these criteria. The cause of this problem is rooted in its learning objective, which uses prediction alone as guidance in search of explanations. A detailed analysis will be provided in the next section.

\section{Analyze Spurious Explanations}~\label{sec:analysis}
\begin{figure}[t]
    \centering

  \subfigure[SCM]{    \label{fig:SEM}
		\includegraphics[width=0.2\textwidth]{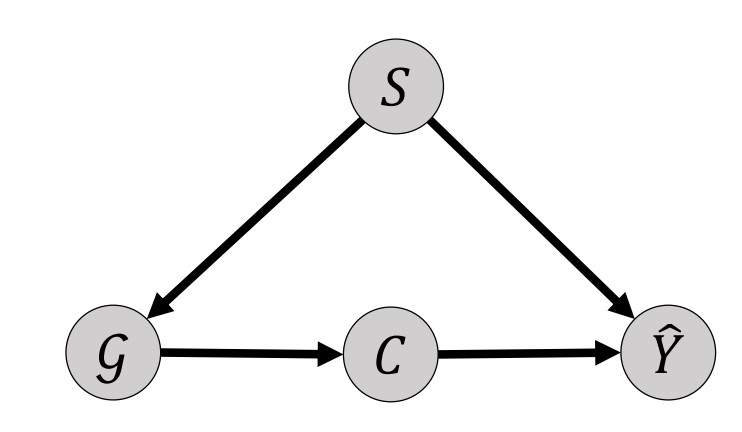}}
  \subfigure[Alignment]{\label{fig:alignment}
		\includegraphics[width=0.18\textwidth]{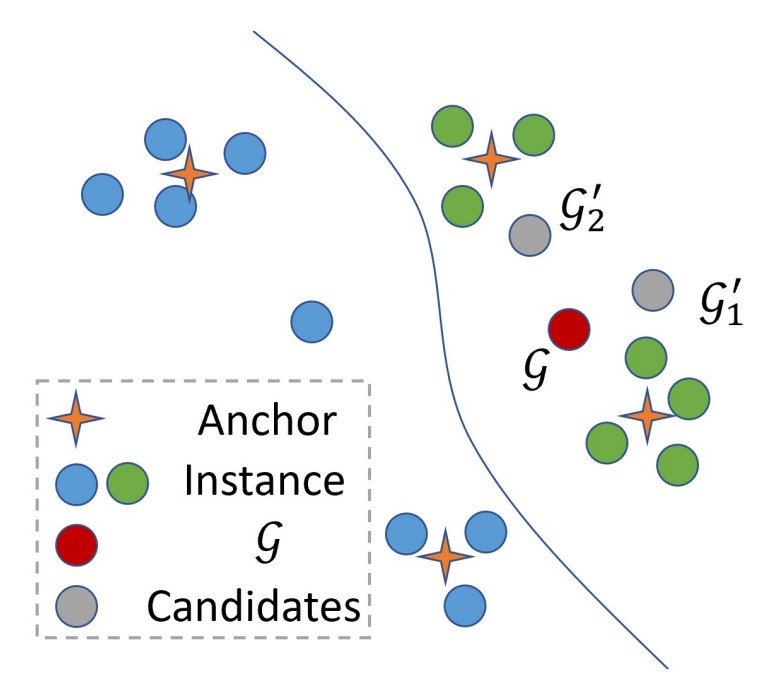}}
    \vskip -2em
    \caption{(a) Prediction rules of $f$, in the form of SCM. (b) An example of anchor-based embedding alignment.}

    \vskip -2em
\end{figure}

With ``spurious explanations'', we refer to those explanations lies outside the genuine rationale of prediction on $\mathcal{G}$, making the usage of $\mathcal{G}'$ as explanations anecdotal. 
As examples in Figure~\ref{fig:case_study}, despite rapid developments in explaining GNNs, the problem w.r.t faithfulness and consistency of detected explanations remains. To get a deeper understanding of reasons behind this problem, we can examine behavior of target GNN model from the causality perspective. Figure ~\ref{fig:SEM} shows the Structural Equation Model (SEM), where variable $C$ denotes discriminative causal factors and variable $S$ represents confounding environment factors. Two paths between $\mathcal{G}$ and the predicted label $\hat{Y}$ can be found.
\begin{itemize}[leftmargin=*]
    \item $\mathcal{G} \rightarrow C \rightarrow \hat{Y}$: This path presents the inference of target GNN model, i.e., critical patterns $C$ that are informative and discriminative for the prediction would be extracted from input graph, upon which the target model is dependent. Causal variables are determined by both the input graph and learned knowledge by the target GNN model. 
    \item $\mathcal{G} \leftarrow S \rightarrow \hat{Y}$: We denote $S$ as the confounding factors, such as depicting the overall distribution of graphs. It is causally related to both the appearance of input graphs and prediction of target GNN models. Masked version of $\mathcal{G}$ could create out-of-distribution (OOD) examples, resulting in spurious causality to prediction outputs. For example in the chemical domain, removing edges (bonds) or nodes (atoms) may obtain invalid molecular graphs that never appear during training. In the existence of distribution shifts, model predictions would be less reliable.
\end{itemize}

Figure~\ref{fig:SEM} provides us with a tool to analyze $f$'s behaviors. From obtained causal structures, we can observe that spurious explanations may arise as a result of failure in recovering original causal rationale. $\mathcal{G}'$ learned from Equation~\ref{eq:framework} may preserve prediction $\hat{Y}$ due to confounding effect of distribution shift or different causal variables $C$ compared to original $\mathcal{G}$. Weakly-trained GNN $f(\cdot)$ that are unstable or non-robust towards noises would further amplify this problem as the prediction is unreliable. 

To further understand the issue, we can build the correspondence from SEM in Figure~\ref{fig:SEM} to the inference process of GNN $f$. Decomposing $f()$ as feature extractor $f_{ext}()$ and classifier $f_{cls}()$, its inference can be summarized as two steps: (1) encoding step with $f_{ext}()$, which takes $\mathcal{G}$ as input and produce its embedding in the representation space $E_C$; (2) classification step with $f_{cls}()$,  which predicts output labels on input's embedding. Connecting these inference steps to SEM in Figure~\ref{fig:SEM}, we can find that: 
\begin{itemize}[leftmargin=*]
    \item The causal path $\mathcal{G} \rightarrow C \rightarrow \hat{Y} $ lies behind the inference process with representation space $E_C$ to encode critical variables $C$;
    \item The confounding effect of distribution shift $S$ works on the inference process via influencing distribution of graph embedding in $E_C$. When masked input $\mathcal{G}'$ is OOD, its embedding would fail to reflect its discriminative features and deviate from real distributions, hence deviating the classification step on it.
\end{itemize}
To summarize, we can observe that spurious explanations are usually obtained due to the following two reasons:
\begin{enumerate}[leftmargin=*]
    \item The obtained $\mathcal{G}'$ is OOD graph. During inference of target GNN model, the encoded representation of $\mathcal{G}'$ is distant from those seen in the training set, making the prediction unreliable;
    \item The encoded discriminative representation does not accord with that of the original graph. Different causal factors ($C$) are extracted between $\mathcal{G}'$ and $\mathcal{G}$, resulting in false explanations. 
\end{enumerate}

\section{Methodology}
Based on the discussion above, in this section, we focus on improving the faithfulness and consistency of GNN explanations and correcting the inductive bias caused by simply relying on prediction outputs. 
We first provide an intuitive introduction to the proposed countermeasure, which takes the internal inference process into account. We then design two concrete algorithms to align $\mathcal{G}$ and $\mathcal{G}'$ in the latent space, to promote that they are seen and processed in the same manner. Finally, theoretical analysis is provided to justify our strategies.

\subsection{Alleviate Spurious Explanations}
Instance-level post-hoc explanation dedicates to finding discriminative substructures that the target model $f$ depends upon. The traditional objective in Equation~\ref{eq:surrogate} can identify minimal predictive parts of input, however, it is dangerous to directly take them as explanations. Due to diversity in graph topology and combinatory nature of sub-graphs, multiple distinct substructures could be identified leading to the same prediction, as discussed in Section~\ref{sec:analysis}.


For an explanation substructure $\mathcal{G}'$ to be faithful, it should follow the same rationale as the original graph $\mathcal{G}$ inside the internal inference of to-be-explained model $f$. To achieve this goal, the explanation $\mathcal{G}'$ should be aligned to $\mathcal{G}$ w.r.t the decision mechanism, reflected in Figure~\ref{fig:SEM}. However, it is non-trivial to extract and compare the critical causal variables $C$ and confounding variables $S$ due to the black box nature of the target GNN model to be explained.

Following the causal analysis in Section~\ref{sec:analysis}, we propose to take an alternative approach by looking into internal embeddings learned by $f$. Causal variables $C$ are encoded in representation space extracted by $f$, and out-of-distribution effects can also be reflected by analyzing embedding distributions. 
An assumption can be safely made: \textit{if two graphs are mapped to embeddings near each other by a GNN layer, then these graphs are seen as similar by it and would be processed similarly by following layers}. 
With this assumption, a proxy task can be designed by aligning internal graph embeddings between $\mathcal{G}'$ and $\mathcal{G}$. This new task can be incorporated into Framework~\ref{eq:framework} as an auxiliary optimization objective.

Let $\mathbf{h}_v^l$ be the representation of node $v$ at the $l$-th GNN layer with $\mathbf{h}_v^0 = \mathbf{F}[v,:]$. Generally, the inference process inside GNN layers can be summarized as a message-passing framework:
    \begin{equation}\label{eq:messagepass}
        \begin{aligned}
        \mathbf{m}_{v}^{l+1} &=\sum_{u \in \mathcal{N}(v)} \text{Message}_{l}\large( \mathbf{h}_v^l, \mathbf{h}_u^l, A_{v,u} \large), \\
        \mathbf{h}_v^{l+1} &= \text{Update}_{l}\large(\mathbf{h}_v^{l}, \mathbf{m}_{v}^{l+1} \large),
        \end{aligned}
    \end{equation}
where $\text{Message}_{l}$ and $\text{Update}_{l}$ are the message function and update function at $l$-th layer, respectively. $\mathcal{N}(v)$ is the set of node $v$'s neighbors. Without loss of generality, the graph pooling layer can also be presented as:
    \begin{equation}\label{eq:gpool}
        \mathbf{h}_{v'}^{l+1} = \sum_{v\in \mathcal{V}} {P}_{v, v'} \cdot \mathbf{h}_{v}^{l}.
    \end{equation}
where ${P}_{v,v'}$ denotes mapping weight from node $v$ in layer $l$ to node $v'$ in layer $l+1$ inside the myriad of GNN for graph classification. We propose to align embedding $\mathbf{h}_v^{l+1}$ at each layer, which contains both node and local neighborhood information.

\subsection{Distribution-Aware Alignment}\label{sec:implement}

Achieving alignment in the embedding space is not straightforward. It has several distinct difficulties. (1) It is difficult to evaluate the distance between $\mathcal{G}$ and $\mathcal{G}'$ in this embedding space. Different dimensions could encode different features and carry different importance. Furthermore, $\mathcal{G}'$ is a substructure of the original $\mathcal{G}$ and a shift on unimportant dimensions would naturally exist. (2) Due to the complexity of graph/node distributions, it is non-trivial to design a measurement of alignments that is both computation-friendly and can correlate well to distance on the distribution manifold. 

To address these challenges, we design a strategy to identify explanatory substructures and preserve their alignment with original graphs in a distribution-aware manner. The basic idea is to utilize other graphs to obtain a global view of the distribution density of embeddings, providing a better measurement of alignment. Concretely, we obtain representative node/graph embeddings as anchors and use distances to these anchors as the distribution-wise representation of graphs. Alignment is conducted on obtained representation of graph pairs. Next, we go into details of this strategy.
    \begin{itemize}[leftmargin=0.2in]
        \item  First, using graphs $\{\mathcal{G}_i\}_{i=1}^{m}$ from the same dataset, a set of node embeddings can be obtained as $\{\{\mathbf{h}^{l}_{v,i}\}_{v \in \mathcal{V}'_i}\}_{i=1}^{m}$ for each layer $l$, where $\mathbf{h}_{v,i}$ denotes embedding of node $v$ in graph $\mathcal{G}_i$. For node-level tasks, we set $\mathcal{V}'_i$ to contain only the center node of graph $\mathcal{G}_i$. For graph-level tasks, $\mathcal{V}'_i$ contains nodes set after graph pooling layer, and we process them following $\{ \sum_{v \in \mathcal{V}'_i} \mathbf{h}^{l+1}_{v,i}/|\mathcal{V}'_i| \}_{i=1}^{m}$ to get global graph representation.
        \item Then, a clustering algorithm is applied to obtained embedding set to get $K$ groups. Clustering centers of these groups are set to be anchors, annotated as $\{\mathbf{h}^{l+1,k}\}_{k=1}^{K}$. In experiments, we select DBSCAN~\cite{ester1996density} as the clustering algorithm, and tune its hyper-parameters to get around $20$ groups.
        \item At $l$-th layer, $\mathbf{h}_v^{l+1}$ is represented in terms of relative distances to those $K$ anchors, as $\mathbf{s}_v^{l} \in \mathbb{R}^{1 \times K}$ with the $k$-th element calculated as $\mathbf{s}_{v,k}^l = \|\mathbf{h}^{l+1}_{v}-\mathbf{h}^{l+1,k}_{v}\|_2$.
    \end{itemize}
Alignment between $\mathcal{G}'$ and $\mathcal{G}$ can be achieved by comparing their representations at each layer. The alignment loss is computed as:
    \begin{equation}
       \mathcal{L}_{align}\large( f(\mathcal{G}), f(\mathcal{G}') \large) = \sum_l  \sum_{v\in \mathcal{V}'} \| \mathbf{s}_{v}^l - \hat{\mathbf{s}}_{v}^l \|_2^2.
    \end{equation}
This metric provides a lightweight strategy for evaluating alignments in the embedding distribution manifold, by comparing relative positions w.r.t representative clustering centers. This strategy can naturally encode the varying importance of each dimension. 
Fig.~\ref{fig:alignment} gives an example, where $\mathcal{G}$ is the graph to be explained and the red stars are anchors. $\mathcal{G}'_1$ and $\mathcal{G}'_2$ are both similar to $\mathcal{G}$ w.r.t absolute distances; while it is easy to see $\mathcal{G}'_1$ is more similar to $\mathcal{G}$ w.r.t to the anchors. In other words, the anchors can better measure the alignment to filter out spurious explanations.

This alignment loss is used as an auxiliary task incorporated into MMI-based framework in Equation~\ref{eq:surrogate} to get faithful explanation as:
\begin{equation}
    \begin{aligned}
    \min_{\mathcal{G}'}  &H_C\big(\hat{Y}, P(\hat{Y}' \mid \mathcal{G}') \big) + \lambda \cdot \mathcal{L}_{Align}, \\
     \quad \text{s.t.} \quad \mathcal{G}' \sim & \mathcal{P}(\mathcal{G}, \mathbf{M}_{A},  \mathbf{M}_{F}), \quad \mathcal{R}(\mathbf{M}_{F},\mathbf{M}_{A}) \leq c 
    \end{aligned}\label{eq:target}
\end{equation}
where $\lambda$ controls the balance between prediction preservation and embedding alignment.  $\mathcal{L}_{Align}$ is flexible to be incorporated into various existing explanation methods.

As a simpler and more direct implementation, we also design a variant based on absolute distance. For layers without graph pooling, the objective can be written as $\sum_l\sum_{v\in \mathcal{V}} \|\mathbf{h}_{v}^l - \hat{\mathbf{h}}_{v}^l \|_2^2 $. For layers with graph pooling, as the structure could be different, we conduct alignment on global representation $ \sum_{v \in \mathcal{V}'} \mathbf{h}^{l+1}_{v}/|\mathcal{V}'| $, where $\mathcal{V}'$ denotes node set after pooling.

\subsection{Theoretical Analysis}

\subsubsection{New Explanation Objective}

\begin{figure}[t!]
  \centering
  \subfigure[Previous Objective]{\label{fig:venn_a}
		\includegraphics[width=0.22\textwidth]{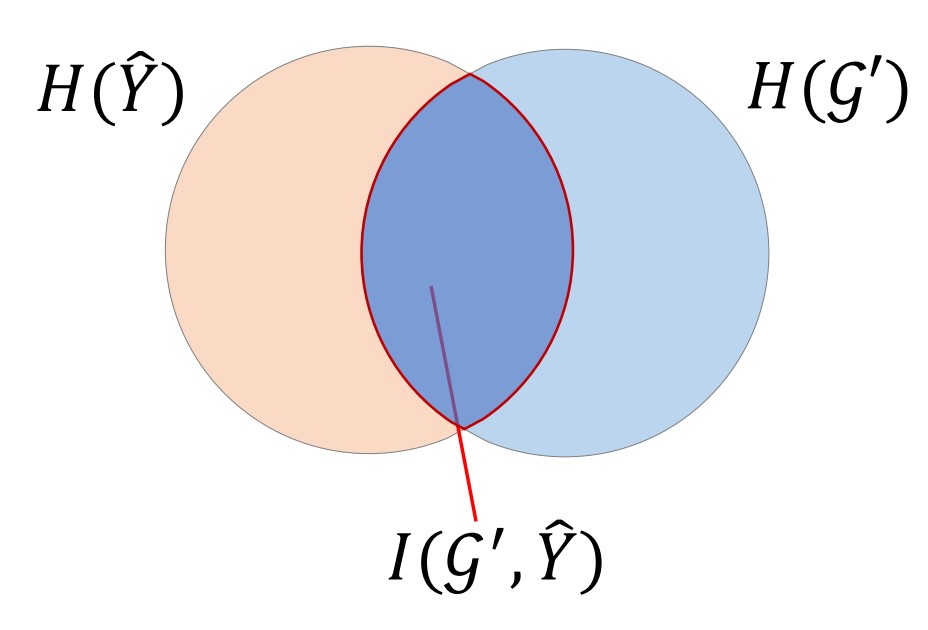}}
  \subfigure[Our Proposed New Objective]{\label{fig:venn_b}
		\includegraphics[width=0.22\textwidth]{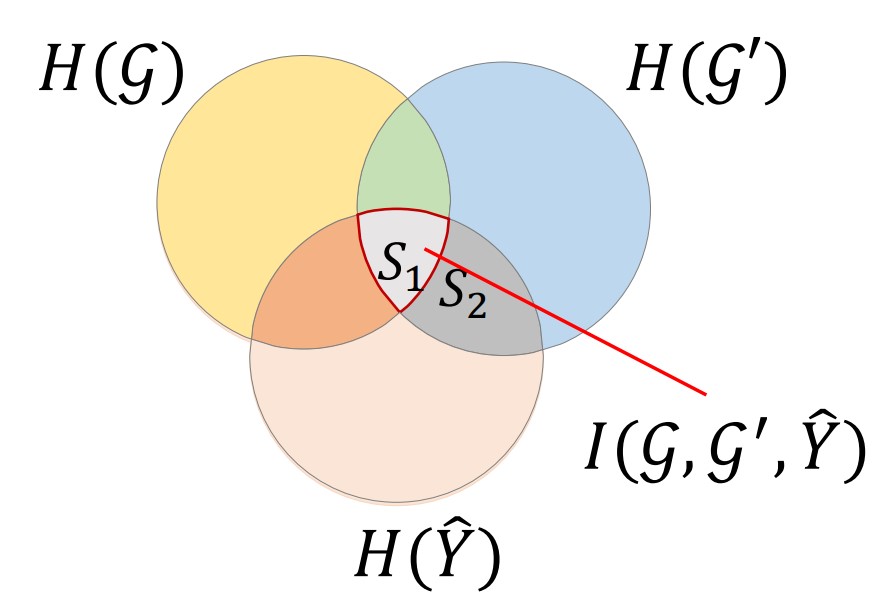}}
    \vskip -1.5em
    \caption{Illustration of our proposed new objective.} \label{fig:venn}
    \vskip -2em
\end{figure}

From previous discussions, it is shown that $\mathcal{G}'$ obtained via Equation \ref{eq:framework} cannot be safely used as explanations. One main drawback of existing GNN explanation methods lies in the inductive bias that the same outcomes do not guarantee the same causes, leaving existing approaches vulnerable towards spurious explanations.  An illustration is given in Figure~\ref{fig:venn}. Objective proposed in Equation \ref{eq:framework} optimizes the mutual information between explanation candidate $\mathcal{G}'$ and $\hat{Y}$, corresponding to maximize the overlapping between $H(\mathcal{G}')$ and $H(\hat{Y})$ in Figure~\ref{fig:venn_a}, or region $S_1 \cup S_2$ in Figure~\ref{fig:venn_b}. Here, $H$ denotes information entropy. However, this learning target cannot prevent the danger of generating spurious explanations. Provided $\mathcal{G}'$ may fall into the region $S_2$, which cannot faithfully represent graph $\mathcal{G}$. Instead, a more sensible objective should be maximizing region $S_1$ in Figure~\ref{fig:venn_b}. The intuition behind this is that in the search input space that causes the same outcome, identified $\mathcal{G}'$ should account for both representative and discriminative parts of original $\mathcal{G}$, to prevent spurious explanations that produce the same outcomes due to different causes. Concretely, finding $\mathcal{G}'$ that maximize $S_1$ can be formalized as:
\begin{equation}~\label{eq:new_frame}
    \begin{aligned}
    \min_{\mathcal{G}'} &- I(\mathcal{G}, \mathcal{G'}, \hat{Y}), \\
    \quad \text{s.t.} \quad \mathcal{G}' \sim & \mathcal{P}(\mathcal{G}, \mathbf{M}_{A},  \mathbf{M}_{F}) \quad \mathcal{R}(\mathbf{M}_{F},\mathbf{M}_{A}) \leq c 
    \end{aligned}
\end{equation}

\subsubsection{Connecting to Our Method}
$I(\mathcal{G}, \mathcal{G'}, \hat{Y})$ is intractable as the latent generation mechanism of $\mathcal{G}$ is unknown. In this part, we expand this objective, connect it to Equation~\ref{eq:target}, and construct its proxy optimizable form as:
\begin{equation} \small
    \begin{aligned}
    & I(\mathcal{G}, \mathcal{G'}, \hat{Y}) = \sum_{y\sim\hat{Y}}\sum_{\mathcal{G}}\sum_{\mathcal{G}'}P(\mathcal{G},\mathcal{G}',y)\cdot \log\frac{P(\mathcal{G}',y)P(\mathcal{G},\mathcal{G}')P(\mathcal{G},y)}{P(\mathcal{G},\mathcal{G}',y)P(\mathcal{G})P(\mathcal{G}')P(y)} \\
    =& \sum_{y\sim\hat{Y}}\sum_{\mathcal{G}}\sum_{\mathcal{G}'}P(\mathcal{G},\mathcal{G}',y)\cdot \log\mathlarger[\frac{P(\mathcal{G}',y)}{P(\mathcal{G}')P(y)}\cdot \frac{P(\mathcal{G},\mathcal{G}')}{P(\mathcal{G})P(\mathcal{G}')} \cdot \frac{P(\mathcal{G},y)}{P(\mathcal{G},y|\mathcal{G}')}\mathlarger] \\
    =& \sum_{y\sim\hat{Y}}\sum_{\mathcal{G}'}P(\mathcal{G}',y)\cdot \log \frac{P(\mathcal{G}',y)}{P(\mathcal{G}')P(y)} +\sum_{\mathcal{G}}\sum_{\mathcal{G}'}P(\mathcal{G},\mathcal{G}')\cdot \log \frac{P(\mathcal{G},\mathcal{G}')}{P(\mathcal{G})P(\mathcal{G}')}\\
    & - \sum_{\mathcal{G}'}\sum_{y\sim\hat{Y}}\sum_{\mathcal{G}}P(\mathcal{G},y,\mathcal{G}')\cdot \log \frac{P(\mathcal{G},y, \mathcal{G}')}{P(\mathcal{G},y)P(\mathcal{G}')}] \\
    =& I(\mathcal{G}', \hat{Y}) + I(\mathcal{G}, \mathcal{G}') - \sum_{y\sim\hat{Y}}\sum_{\mathcal{G}}P(\mathcal{G},y)\sum_{\mathcal{G}'}P(\mathcal{G}'|\mathcal{G},y)\cdot \log P(\mathcal{G}'|\mathcal{G},y) \\
    &+ \sum_{\mathcal{G}'}\sum_{y\sim\hat{Y}}\sum_{\mathcal{G}}P(\mathcal{G},y,\mathcal{G}')\cdot \log P(\mathcal{G}') \\
    =& I(\mathcal{G}', \hat{Y}) + I(\mathcal{G}, \mathcal{G}') + H(\mathcal{G}' | \mathcal{G},\hat{Y})-H(\mathcal{G}').
    \nonumber
    \end{aligned}
\end{equation}
Since both $H(\mathcal{G}' | \mathcal{G},\hat{Y})$ and $H(\mathcal{G}')$ depicts entropy of explanation $\mathcal{G}'$ and are closely related to perturbation budgets, we can neglect these two terms and get a surrogate optimization objective for $\max_{\mathcal{G}'}I(\mathcal{G}, \mathcal{G'}, \hat{Y}) $ as $\max_{\mathcal{G}'}I(\hat{Y},\mathcal{G'})+I(\mathcal{G}',\mathcal{G})$.

In $\max_{\mathcal{G}'}I(\hat{Y},\mathcal{G'})+I(\mathcal{G}',\mathcal{G})$, the first term $\max_{\mathcal{G}'}I(\hat{Y},\mathcal{G'})$ is the same as Eq.( ~\ref{eq:framework}). Following ~\cite{ying2019gnnexplainer}, We relax it as $\min_{\mathcal{G}'}H_C(\hat{Y}, \hat{Y}'|\mathcal{G'})$, optimizing $\mathcal{G}'$ to preserve original prediction outputs. The second term, $\max_{\mathcal{G}'}I(\mathcal{G}',\mathcal{G})$, corresponds to maximizing consistency between $\mathcal{G}'$ and $\mathcal{G}$. Although the graph generation process is latent, with the safe assumption that embedding $\mathbf{E}_{\mathcal{G}}$ extracted by $f$ is representative of $\mathcal{G}$, we can construct a proxy objective $\max_{\mathcal{G}'}I(\mathbf{E}_{\mathcal{G}'},\mathbf{E}_\mathcal{G})$, improving the consistency in the embedding space. In this work, we optimize this objective by aligning their representations, either optimizing a simplified distance metric or conducting distribution-aware alignment.

\section{Experiment}

In this section, we conduct a set of experiments to evaluate the benefits of the proposed auxiliary task in providing instance-level post-hoc explanations. Experiments are conducted on $5$ datasets, and obtained explanations are evaluated w.r.t both faithfulness and consistency. Particularly, we aim to answer the following questions:
\begin{itemize}[leftmargin=0.2in]
    \item \textbf{RQ1} Can the proposed framework perform strongly in identifying explanatory sub-structures for interpreting GNNs?
    \item \textbf{RQ2} Is the consistency problem severe in existing GNN explanation methods? Could the proposed embedding alignment improve GNN explainers over this criterion?
    \item \textbf{RQ3} Can our proposed strategy prevent spurious explanations and be more faithful to the target GNN model?
\end{itemize}

\subsection{Experiment Settings}
\subsubsection{Datasets}
We conduct experiments on five publicly available benchmark datasets for explainability of GNNs:

\begin{itemize}[leftmargin=*]
    \item BA-Shapes\cite{ying2019gnnexplainer}: A node classification dataset with a Barabasi-Albert (BA) graph of $300$ nodes as the base structure. $80$ ``house” motifs are randomly attached to the base graph. Nodes are labeled based on positions.  Edges inside the corresponding motif are ground-truth for explaining 
those attached nodes.
    \item Tree-Grid ~\cite{ying2019gnnexplainer}: A node classification dataset created by attaching $80$ grid motifs to a single $8$-layer balanced binary tree. For nodes in the attached motif, edges inside the same motif are used as ground-truth explanations.
    \item Infection~\cite{faber2021comparing}: A single network initialized with an ER random graph. $5\%$ of nodes are labeled as infected, and other nodes are labeled based on their shortest distances to those infected ones. The graph is pre-processed following~\cite{faber2021comparing}. For each node, its shortest path is used as the ground-truth explanation.
    \item Mutag~\cite{ying2019gnnexplainer}: A graph classification dataset. Each graph corresponds to a molecule with nodes for atoms and edges for chemical bonds. Chemical groups are identified as explanations using prior domain knowledge. Following PGExplainer~\cite{luo2020parameterized}, chemical groups \textit{$NH_2$} and \textit{$NO_2$} are used as ground-truth explanations.
    \item Graph-SST5~\cite{yuan2020explainability}: A graph classification dataset constructed from text data, with labels from sentiment analysis. In this dataset, there is no ground-truth explanation provided, and heuristic metrics are usually adopted for evaluation.
\end{itemize}

\subsubsection{Configurations}
Following existing work~\cite{luo2020parameterized}, a three-layer GCN~\cite{kipf2016semi} is trained on $80\%$ instances of each dataset as the target model. All explainers are trained using ADAM optimizer with weight decay set to $5e$-$4$. For GNNExplainer, learning rate is initialized to $0.01$ with training epoch being $100$. For PGExplainer, learning rate is initialized to $0.003$ and training epoch is set as $30$. Hyper-parameter $\lambda$, which controls the weight of $\mathcal{L}_{align}$, is tuned via grid search. Explanations are tested on all instances.

To evaluate the effectiveness of the proposed framework,  we select a group of representative instance-level post-hoc GNN explanation methods as baselines:  GRAD~\cite{luo2020parameterized}, ATT~\cite{luo2020parameterized}, GNNExplainer~\cite{ying2019gnnexplainer}, PGExplaienr~\cite{luo2020parameterized},  Gem~\cite{lin2021generative} and RG-Explainer~\cite{shan2021reinforcement}. Our proposed algorithms in Section~\ref{sec:implement} are implemented and incorporated into two representative GNN explanation frameworks, i.e., GNNExplainer~\cite{ying2019gnnexplainer} and PGExplainer~\cite{luo2020parameterized}. 

\subsubsection{Evaluation Metrics}
To evaluate \textit{faithfulness} of different methods, following~\cite{yuan2020explainability}, we adopt two metrics: (1) AUROC score on edge importance and (2) Fidelity of explanations. On benchmarks with oracle explanations available, we can compute the AUROC score on identified edges as the well-trained target GNN should follow those predefined explanations. On datasets without ground-truth explanations, we evaluate explanation quality with fidelity measurement following~\cite{yuan2020explainability}. Concretely, we observe prediction changes by sequentially removing edges following assigned importance weight, and a faster performance drop represents stronger fidelity. 

To evaluate \textit{consistency} of explanations, we randomly run each method $5$ times, and report average structural hamming distance (SHD)~\cite{tsamardinos2006max} among obtained explanations. A smaller SHD  score indicates stronger consistency.

\begin{table}[t!]
  \setlength{\tabcolsep}{4.5pt}
  
  \caption{Explanation Faithfulness in terms of AUC on Edges}\label{tab:expl_auroc} 
  \vskip -1em
  \begin{tabular}{p{2.1cm} | P{1.23cm}  P{1.23cm}   P{1.23cm}  P{1.23cm} }

    \hline
     & BA-Shapes &  Tree-Grid  & Infection & Mutag \\
    \hline

    \hline
    GRAD & $88.2$ & $61.2$  & $74.0$ & $78.3$ \\
    ATT & $81.5$ & $66.7$  & -- & $76.5$ \\
    Gem & $97.1$ & --  & -- & $83.4$ \\
    RG-Explainer & $98.5$ & $92.7$ & -- & $87.3$ \\
    \hline
    GNNExplainer & $93.1_{\pm1.8}$ &  $86.2_{\pm2.2}$   & $92.2_{\pm1.1}$ & $74.9_{\pm1.9}$ \\
    + Align\_Emb & $95.3_{\pm1.4}$ &  $91.2_{\pm2.3}$ & $93.0_{\pm1.0}$ & $76.3_{\pm1.7}$ \\
    + Align\_Anchor & $97.1_{\pm1.3}$ & $92.4_{\pm1.9}$ & $\textbf{93.1}_{\pm0.8}$ & $78.9_{\pm1.6}$ \\    
    \hline
    PGExplainer & $96.9_{\pm0.7}$ & $92.7_{\pm1.5}$ &   $89.6_{\pm0.6}$ & $83.7_{\pm1.2}$ \\
    + Align\_Emb & $97.2_{\pm0.7}$ & $\mathbf{95.8}_{\pm0.9}$ &   $90.5_{\pm0.7}$ & $92.8_{\pm1.1}$ \\ 
    + Align\_Anchor & $\mathbf{98.7}_{\pm0.5}$ & $94.7_{\pm1.2}$   & $91.6_{\pm0.6}$ & $\mathbf{94.5}_{\pm0.8}$ \\
    \hline
  \end{tabular}
  \vskip -1em
\end{table}

\subsection{Explanation Faithfulness}
To answer \textbf{RQ1}, we compare explanation methods in terms of AUROC score and explanation fidelity.

\subsubsection{AUROC on Edges}
In this subsection, AUROC scores of different methods are reported by comparing assigned edge importance weight with ground-truth explanations. For baseline methods GRAD, ATT, Gem, and RG-Explainer, their performances reported in their original papers are presented. GNNExplainer and PGExplainer are re-implemented, upon which two alignment strategies are instantiated and tested. Each experiment is randomly conducted $5$ times, and we summarize the average performance in Table~\ref{tab:expl_auroc}. A higher AUROC score indicates more accurate explanations. From the results, we can make the following observations:
\begin{itemize}[leftmargin=*]
    \item Across all four datasets, with both GNNExplainer or PGExplainer as the base method, incorporating embedding alignment can improve the quality of obtained explanations;
    \item In most cases, anchor-based alignment demonstrates stronger improvements, which shows the benefits of utilizing global distribution information;
    \item On more complex datasets like Mutag, the benefit of introducing embedding alignment is more significant, e.g., the performance of PGExplainer improves from 83.7\% to 94.5\% with Align Anchor. This result indicates that the problem of spurious explanations is severer with increased dataset complexity.   
\end{itemize}

\subsubsection{Explanation Fidelity}
\begin{figure}[t!]
  \centering
  \subfigure[Tree-Grid, GNNExplainer]{
		\includegraphics[width=0.2\textwidth]{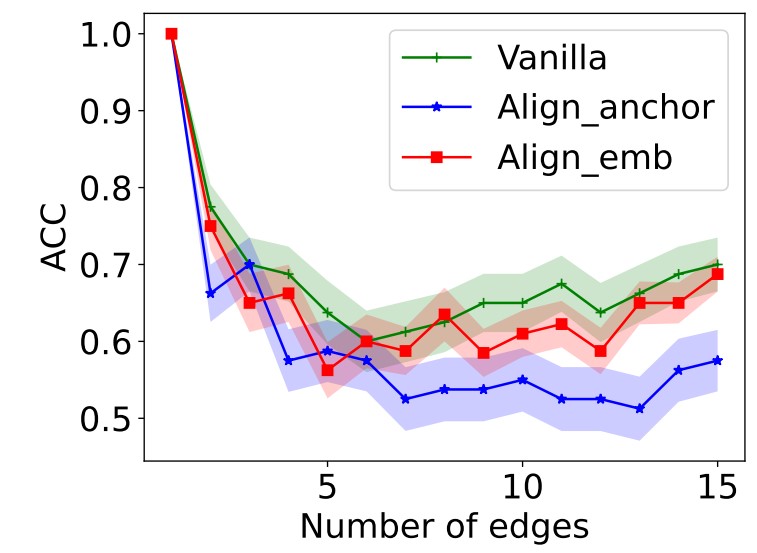}}
  \subfigure[Tree-Grid, PGExplainer]{
		\includegraphics[width=0.2\textwidth]{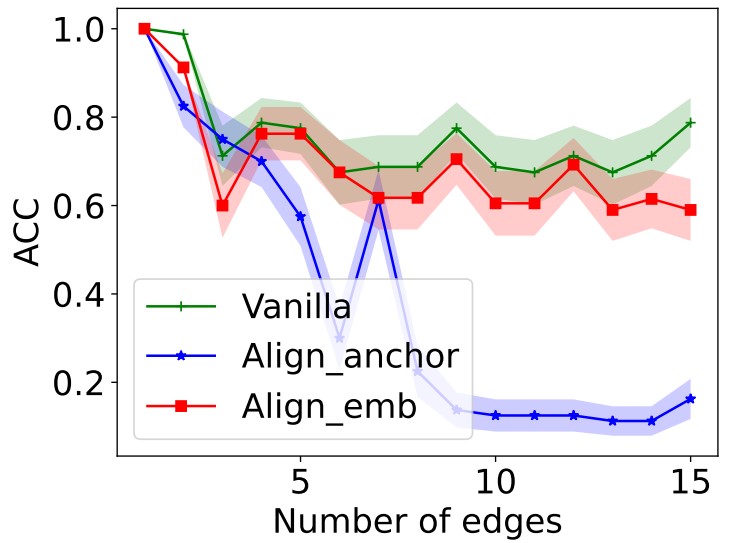}}
  \subfigure[Graph-SS5,  GNNExplainer]{
		\includegraphics[width=0.2\textwidth]{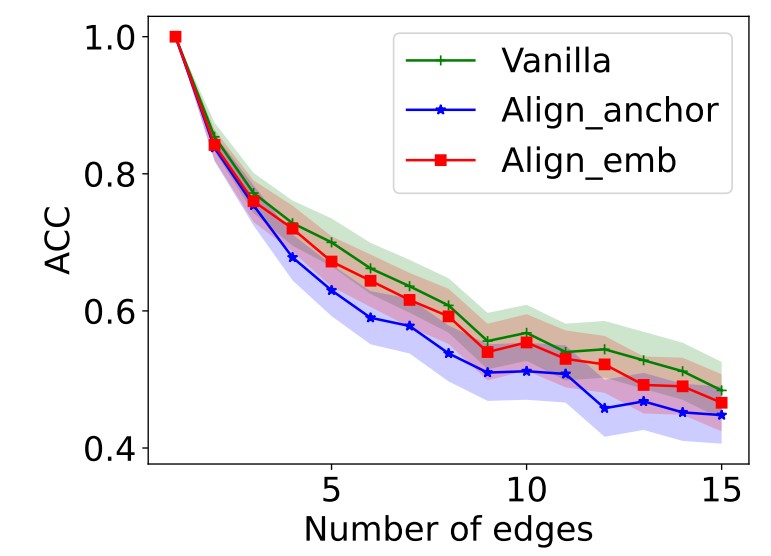}}
  \subfigure[Graph-SS5,  PGExplainer]{
		\includegraphics[width=0.2\textwidth]{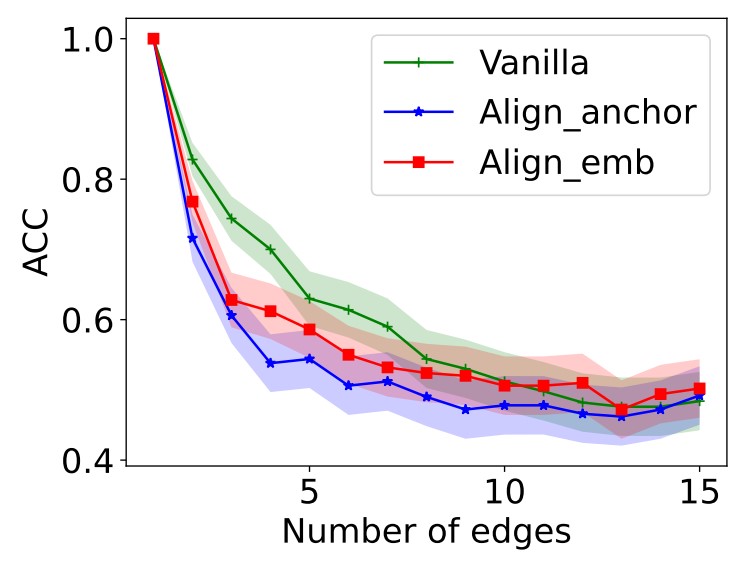}}
    \vskip -1.5em
    \caption{Explanation Fidelity.} \label{fig:fidelity}
    \vskip -2em
\end{figure}

In addition to comparing to ground-truths explanations, we also evaluate the obtained explanations in terms of fidelity. Specifically, we sequentially
remove edges from the graph by following importance weight learned by the explanation model and test the classification performance. Generally, the removal of really important edges would significantly degrade the classification performance. Thus, a faster performance drop represents stronger fidelity. We conduct experiments on Tree-Grid and Graph-SST5. Each experiment is conducted $3$ times and we report results averaged across all instances on each dataset. PGExplainer and GNNExplainer are used as the backbone method. We plot the curves of prediction accuracy concerning the number of removed edges in  Fig.~\ref{fig:fidelity}. From the figure, we can observe that when proposed embedding alignment is incorporated, the classification accuracy from edge removal drops much faster, which shows that the proposed embedding alignment can help to identify more important edges used by GNN for classification, hence providing better explanations. Furthermore, anchor-based alignment is shown to be more effective, validating advantages in distribution-aware alignment. 

From these two experiments, we can observe that embedding alignment can obtain explanations of better faithfulness and is flexible to be incorporated into various models such as GNNExplainer and PGExplainer, which answers RQ1.

\subsection{Explanation Consistency}
One problem of spurious explanation is that, due to the randomness in initialization of the explainer, the explanation for the same instance given by a GNN explainer could be different for different runs, which violates the \textit{consistency} of explanations. To test the severity of this problem and answer \textbf{RQ2}, we evaluate the proposed framework in terms of explanation consistency. We adopt GNNExplainer and PGExplainer as baselines. Specifically, SHD distance among explanatory edges with top-$k$ importance weights identified each time is computed. Then, the results are averaged for all instances in the test set. Each experiment is conducted $5$ times, and the average consistency on dataset Tree-Grid and Mutag are reported in Table~\ref{tab:consistency_Tree} and Table~\ref{tab:consistency_Mutag}, respectively. Larger distances indicate inconsistent explanations. From the table, we can observe that existing method following Equation~\ref{eq:framework} suffers from the consistency problem. For example, average SHD distance on top-$6$ edges is $4.39$ for GNNExplainer. Introducing the auxiliary task of aligning embeddings can significantly improve explainers in terms of this criterion. After incorporating anchor-based alignment on dataset TreeGrid, SHD distance of top-$6$ edges drops from $4.39$ to $2.21$ for GNNExplainer and from $1.38$ to $0.13$ for PGExplainer, which validates effectiveness of our proposal in obtaining consistent explanations.

\begin{table}[t!]
  \setlength{\tabcolsep}{4.5pt}
  
  \caption{Consistency of explanation in terms of average SHD distance across $5$ rounds of random running on Tree-Grid.}\label{tab:consistency_Tree} 
  \vskip -1em
  \begin{tabular}{p{1.9cm} |  P{0.6cm}  P{0.6cm}  P{0.6cm}  P{0.6cm}  P{0.6cm}  P{0.6cm}}

    \hline
     &  \multicolumn{6}{c}{Top-K Edges}  \\
    \hline
    Methods & 1 & 2 & 3 & 4 & 5 & 6  \\
    \hline
    GNNExplainer & $0.86$ & $1.85$ & $2.48$ & $3.14$ & $3.77$ & $4.39$ \\
    +Align\_Emb & $0.77$ & $1.23$ & $1.28$ & $0.96$ & $1.81$ & $2.72$ \\
     +Align\_Anchor & $0.72$ & $1.06$ & $0.99$ & $0.53$ & $1.52$ & $2.21$ \\
    \hline
    PGExplainer  & $0.74$ & $1.23$ & $0.76$ & $0.46$ & $0.78$ & $1.38$ \\
    +Align\_Emb  & $0.11$ & $0.15$ & $\mathbf{0.13}$ & $\mathbf{0.11}$ & $0.24$ & $0.19$ \\
    +Align\_Anchor  & $\mathbf{0.07}$ & $\mathbf{0.12}$ & $\mathbf{0.13}$ & $0.16$ & $\mathbf{0.21}$ & $\mathbf{0.13}$  \\
    \hline
  \end{tabular}
    \vskip -1em
\end{table}

\begin{table}[t!]
  \setlength{\tabcolsep}{4.5pt}
  
  \caption{Consistency of explanation in terms of average SHD distance across $5$ rounds of random running on Mutag.}\label{tab:consistency_Mutag} 
  \vskip -1em
  \begin{tabular}{p{1.9cm} |  P{0.6cm}  P{0.6cm}  P{0.6cm}  P{0.6cm}  P{0.6cm}  P{0.6cm}}

    \hline
     &  \multicolumn{6}{c}{Top-K Edges}  \\
    \hline
    Methods & 1 & 2 & 3 & 4 & 5 & 6  \\
    \hline
    GNNExplainer & $1.12$ & $1.74$ & $2.65$ & $3.40$ & $4.05$ & $4.78$ \\
    +Align\_Emb & $1.05$ & $1.61$ & $2.33$ & $3.15$ & $3.77$ & $4.12$ \\
     +Align\_Anchor & $1.06$ & $1.59$ & $2.17$ & $3.06$ & $3.54$ & $3.95$ \\
    \hline
    PGExplainer  & $0.91$ & $1.53$ & $2.10$ & $2.57$ & $3.05$ & $3.42$ \\
    +Align\_Emb  & $0.55$ & $0.96$ & $1.13$ & $1.31$ & $1.79$ & $2.04$ \\
    +Align\_Anchor  & $\mathbf{0.51}$ & $\mathbf{0.90}$ & $\mathbf{1.05}$ & $\mathbf{1.27}$ & $\mathbf{1.62}$ & $\mathbf{1.86}$  \\
    \hline
  \end{tabular}
    \vskip -1em
\end{table}

\subsection{Ability in Avoiding Spurious Explanations}
Existing graph explanation benchmarks are usually designed to be less ambiguous, containing only one oracle cause of labels, and identified explanatory substructures are evaluated via comparing with the ground-truth explanation. However, this result could be misleading, as faithfulness of explanation methods in more complex scenarios is left untested. Real-world datasets are usually rich in spurious patterns and a trained GNN could contain diverse biases, setting a tighter requirement on explanation methods. Thus, to evaluate if our framework can alleviate the spurious explanation issue and answer \textbf{RQ3}, we create a new graph-classification dataset: MixMotif, which enables us to train a biased GNN model, and test whether explanation methods can successfully expose this bias. 

Specifically, inspired by ~\cite{wu2022discovering}, we design three types of base graphs, i.e., Tree, Ladder, and Wheel, and three types of motifs, i.e., Cycle, House, and Grid. With a mix ratio $\gamma$, motifs are preferably attached to base graphs. Labels are set as the type of motif. When $\gamma$ is set to $0$, each motif has the same probability of being attached to three base graphs. Thus, we consider the dataset with $\gamma=0$ as clean or bias-free. We would expect GNN trained with $\gamma=0$ to focus on the motif structure. However, when $\gamma$ becomes larger, the spurious correlation between base graph and the label would exist, i.e., a GNN might utilize the base graph structure for motif classification instead of relying on the motif structure. 

In this experiment, we set $\gamma$ to $0$ and $0.7$ separately, and train GNN $f_{0}$ and $f_{0.7}$ for each setting. Two models are tested in graph classification performance. Then, explanation methods are applied to and fine-tuned on $f_0$. Following that, these explanation methods are applied to explain $f_{0.7}$ using found hyper-parameters. Results are summarized in Table~\ref{tab:biasedExpl}.

\begin{table}[t!]
  \setlength{\tabcolsep}{4.5pt}
  
  \caption{Performance on MixMotif. Two GNNs are trained with different $\gamma$. We check their performance in graph classification, then compare obtained explanations with the motif. }
  \vskip -1em
  \begin{tabular}{p{1cm} | p{0.4cm} |  P{1.4cm}  P{1.2cm} | P{1.4cm}  P{1.2cm} }

    \hline
    \multicolumn{2}{c}{} &  \multicolumn{4}{c}{$\gamma$ in Training}  \\
    \hline
    \multicolumn{2}{c|}{Classification} & \multicolumn{2}{c|}{$0$} & \multicolumn{2}{c}{$0.7$} \\
    \hline
    \multirow{2}{*}{$\gamma$ in test} & $0$ & \multicolumn{2}{c|}{$0.982$}  &  \multicolumn{2}{c}{$0.765$} \\
     &$0.7$ & \multicolumn{2}{c|}{$0.978$}  &  \multicolumn{2}{c}{$0.994$} \\
    \hline 
    \multicolumn{2}{c|}{Explanation} & PGExplainer & +Align & PGExplainer & +Align \\
    \hline
    \multicolumn{2}{c|}{\multirow{2}{*}{\parbox{1.4cm}{AUROC on Motif}}} & $0.711$ & $\mathbf{0.795}$ & $0.748$ & $\mathbf{0.266}$ \\
    \multicolumn{2}{c|}{}& \multicolumn{2}{c|}{(Higher is better)} & \multicolumn{2}{c}{(Lower is better)} \\
    \hline
  \end{tabular}\label{tab:biasedExpl}
  \vskip -1em
\end{table}

From Table~\ref{tab:biasedExpl}, we can observe that (1) $f_0$ achieves almost perfect graph classification performance during testing. This high accuracy indicates that it captures the genuine pattern, relying on motifs to make predictions. Looking at explanation results, it is shown that our proposal offers more faithful explanations, achieving higher AUROC on motifs. (2) $f_{0.7}$ fails to predict well with $\gamma=0$, showing that there are biases in it and it no longer depends solely on the motif structure for prediction. Although ground-truth explanations are unknown in this case, a successful explanation should expose this bias. However, PGExplainer would produce similar explanations as the clean model, still highly in accord with motif structures. Instead, for explanations produced by embedding alignment, AUROC score would drop from $0.795$ to $0.266$, exposing the change in prediction rationales, hence able to expose biases. (3) In summary, our proposal can provide more faithful explanations for both clean and mixed settings, while PGExplainer would suffer from spurious explanations and fail to faithfully explain GNN's predictions, especially in the existence of biases.

\subsection{Hyperparameter Sensitivity Analysis}
\begin{figure}[t!]
  \centering
  \subfigure[Tree-Grid]{
		\includegraphics[width=0.22\textwidth]{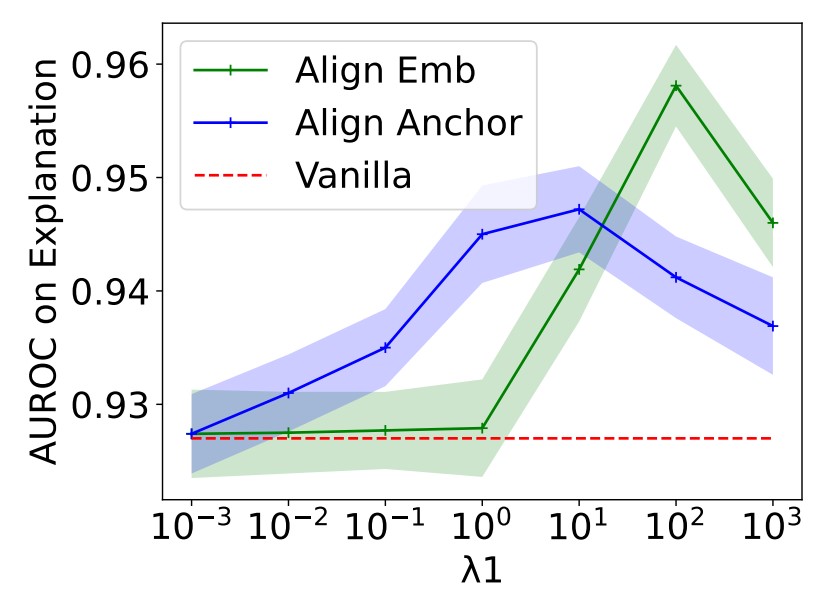}}
  \subfigure[Mutag]{
		\includegraphics[width=0.22\textwidth]{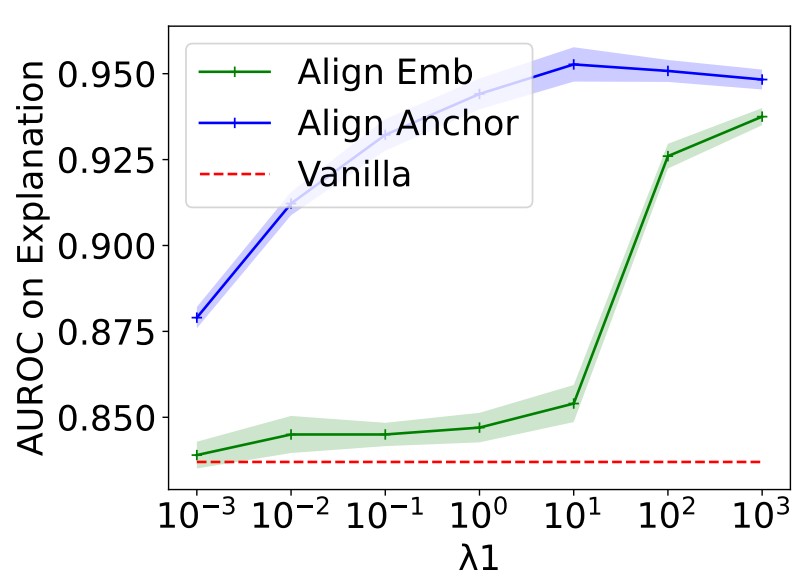}}
    \vskip -1.5em
    \caption{Sensitivity of PGExplainer towards weight of embedding alignment loss. } \label{fig:parameter_sensitivity}
    \vskip -2em
\end{figure}

In this part, we vary the hyper-parameter $\lambda$ to test method's sensitivity towards its values. $\lambda$ controls the weight of our proposed embedding alignment task. To keep simplicity, all other configurations are kept unchanged, and $\lambda$ is varied as scale $[1e-3,1e-2, 1e-1, 1, 10,1e2, 1e3\}$. PGExplainer is adopted as the base method. Each experiment is conducted $3$ times on Tree-Grid and Mutag. Averaged results are visualized in Figure~\ref{fig:parameter_sensitivity}. From the figure, we can observe that increasing $\lambda$ has a positive effect at first, and the further increase would result in a performance drop. Besides, anchor-based alignment usually requires a smaller weight to show improvements.

\section{Conclusion}
In this work, we study a novel problem of obtaining faithful and consistent explanations for GNNs, which is largely neglected by existing MMI-based explanation framework. With close analysis on inference of GNNs, we propose a simple yet effective approach by aligning internal embeddings of the raw graph and its candidate explainable subgraph. The designed algorithms can be incorporated into existing methods with no effort. Experiments validate its effectiveness, and further theoretical analysis shows that it is more faithful in design.
In the future, we will seek more robust explanations. Increased robustness indicates stronger generality, and could provide better class-level interpretation at the same time. 

\section{Acknowledgement}
This material is based upon work supported by, or in part by, the National Science Foundation under grants number IIS-1707548 and IIS-1909702, the Army Research Office under grant number W911NF21-1-0198, and DHS CINA under grant number E205949D. The findings and conclusions in this paper do not necessarily reflect the view of the funding agency.

\bibliographystyle{ACM-Reference-Format}
 \balance 
\bibliography{acmart}


\begin{thebibliography}{44}


\ifx \showCODEN    \undefined \def \showCODEN     #1{\unskip}     \fi
\ifx \showDOI      \undefined \def \showDOI       #1{#1}\fi
\ifx \showISBNx    \undefined \def \showISBNx     #1{\unskip}     \fi
\ifx \showISBNxiii \undefined \def \showISBNxiii  #1{\unskip}     \fi
\ifx \showISSN     \undefined \def \showISSN      #1{\unskip}     \fi
\ifx \showLCCN     \undefined \def \showLCCN      #1{\unskip}     \fi
\ifx \shownote     \undefined \def \shownote      #1{#1}          \fi
\ifx \showarticletitle \undefined \def \showarticletitle #1{#1}   \fi
\ifx \showURL      \undefined \def \showURL       {\relax}        \fi
\providecommand\bibfield[2]{#2}
\providecommand\bibinfo[2]{#2}
\providecommand\natexlab[1]{#1}
\providecommand\showeprint[2][]{arXiv:#2}

\bibitem[\protect\citeauthoryear{Atwood and Towsley}{Atwood and
  Towsley}{2016}]%
        {atwood2016diffusion}
\bibfield{author}{\bibinfo{person}{James Atwood} {and} \bibinfo{person}{Don
  Towsley}.} \bibinfo{year}{2016}\natexlab{}.
\newblock \showarticletitle{Diffusion-convolutional neural networks}. In
  \bibinfo{booktitle}{\emph{Advances in neural information processing
  systems}}. \bibinfo{pages}{1993--2001}.
\newblock


\bibitem[\protect\citeauthoryear{Baldassarre and Azizpour}{Baldassarre and
  Azizpour}{2019}]%
        {baldassarre2019explainability}
\bibfield{author}{\bibinfo{person}{Federico Baldassarre} {and}
  \bibinfo{person}{Hossein Azizpour}.} \bibinfo{year}{2019}\natexlab{}.
\newblock \showarticletitle{Explainability techniques for graph convolutional
  networks}.
\newblock \bibinfo{journal}{\emph{arXiv preprint arXiv:1905.13686}}
  (\bibinfo{year}{2019}).
\newblock


\bibitem[\protect\citeauthoryear{Bruna, Zaremba, Szlam, and LeCun}{Bruna
  et~al\mbox{.}}{2013}]%
        {bruna2013spectral}
\bibfield{author}{\bibinfo{person}{Joan Bruna}, \bibinfo{person}{Wojciech
  Zaremba}, \bibinfo{person}{Arthur Szlam}, {and} \bibinfo{person}{Yann
  LeCun}.} \bibinfo{year}{2013}\natexlab{}.
\newblock \showarticletitle{Spectral networks and locally connected networks on
  graphs}.
\newblock \bibinfo{journal}{\emph{arXiv preprint arXiv:1312.6203}}
  (\bibinfo{year}{2013}).
\newblock


\bibitem[\protect\citeauthoryear{Chereda, Bleckmann, Kramer, Leha, and
  Bei{\ss}barth}{Chereda et~al\mbox{.}}{2019}]%
        {Chereda2019UtilizingMN}
\bibfield{author}{\bibinfo{person}{Hryhorii Chereda}, \bibinfo{person}{A.
  Bleckmann}, \bibinfo{person}{F. Kramer}, \bibinfo{person}{A. Leha}, {and}
  \bibinfo{person}{T. Bei{\ss}barth}.} \bibinfo{year}{2019}\natexlab{}.
\newblock \showarticletitle{Utilizing Molecular Network Information via Graph
  Convolutional Neural Networks to Predict Metastatic Event in Breast Cancer}.
\newblock \bibinfo{journal}{\emph{Studies in health technology and
  informatics}}  \bibinfo{volume}{267} (\bibinfo{year}{2019}),
  \bibinfo{pages}{181--186}.
\newblock


\bibitem[\protect\citeauthoryear{Dai, Jin, Liu, and Wang}{Dai
  et~al\mbox{.}}{2022a}]%
        {dai2022towards}
\bibfield{author}{\bibinfo{person}{Enyan Dai}, \bibinfo{person}{Wei Jin},
  \bibinfo{person}{Hui Liu}, {and} \bibinfo{person}{Suhang Wang}.}
  \bibinfo{year}{2022}\natexlab{a}.
\newblock \showarticletitle{Towards Robust Graph Neural Networks for Noisy
  Graphs with Sparse Labels}.
\newblock \bibinfo{journal}{\emph{arXiv preprint arXiv:2201.00232}}
  (\bibinfo{year}{2022}).
\newblock


\bibitem[\protect\citeauthoryear{Dai and Wang}{Dai and Wang}{2021}]%
        {dai2021towards}
\bibfield{author}{\bibinfo{person}{Enyan Dai} {and} \bibinfo{person}{Suhang
  Wang}.} \bibinfo{year}{2021}\natexlab{}.
\newblock \showarticletitle{Towards Self-Explainable Graph Neural Network}. In
  \bibinfo{booktitle}{\emph{Proceedings of the 30th ACM International
  Conference on Information \& Knowledge Management}}.
  \bibinfo{pages}{302--311}.
\newblock


\bibitem[\protect\citeauthoryear{Dai, Zhao, Zhu, Xu, Guo, Liu, Tang, and
  Wang}{Dai et~al\mbox{.}}{2022b}]%
        {dai2022comprehensive}
\bibfield{author}{\bibinfo{person}{Enyan Dai}, \bibinfo{person}{Tianxiang
  Zhao}, \bibinfo{person}{Huaisheng Zhu}, \bibinfo{person}{Junjie Xu},
  \bibinfo{person}{Zhimeng Guo}, \bibinfo{person}{Hui Liu},
  \bibinfo{person}{Jiliang Tang}, {and} \bibinfo{person}{Suhang Wang}.}
  \bibinfo{year}{2022}\natexlab{b}.
\newblock \showarticletitle{A Comprehensive Survey on Trustworthy Graph Neural
  Networks: Privacy, Robustness, Fairness, and Explainability}.
\newblock \bibinfo{journal}{\emph{arXiv preprint arXiv:2204.08570}}
  (\bibinfo{year}{2022}).
\newblock


\bibitem[\protect\citeauthoryear{Duvenaud, Maclaurin, Iparraguirre, Bombarell,
  Hirzel, Aspuru-Guzik, and Adams}{Duvenaud et~al\mbox{.}}{2015}]%
        {duvenaud2015convolutional}
\bibfield{author}{\bibinfo{person}{David~K Duvenaud}, \bibinfo{person}{Dougal
  Maclaurin}, \bibinfo{person}{Jorge Iparraguirre}, \bibinfo{person}{Rafael
  Bombarell}, \bibinfo{person}{Timothy Hirzel}, \bibinfo{person}{Al{\'a}n
  Aspuru-Guzik}, {and} \bibinfo{person}{Ryan~P Adams}.}
  \bibinfo{year}{2015}\natexlab{}.
\newblock \showarticletitle{Convolutional networks on graphs for learning
  molecular fingerprints}. In \bibinfo{booktitle}{\emph{Advances in neural
  information processing systems}}. \bibinfo{pages}{2224--2232}.
\newblock


\bibitem[\protect\citeauthoryear{Ester, Kriegel, Sander, Xu,
  et~al\mbox{.}}{Ester et~al\mbox{.}}{1996}]%
        {ester1996density}
\bibfield{author}{\bibinfo{person}{Martin Ester}, \bibinfo{person}{Hans-Peter
  Kriegel}, \bibinfo{person}{J{\"o}rg Sander}, \bibinfo{person}{Xiaowei Xu},
  {et~al\mbox{.}}} \bibinfo{year}{1996}\natexlab{}.
\newblock \showarticletitle{A density-based algorithm for discovering clusters
  in large spatial databases with noise.}. In \bibinfo{booktitle}{\emph{kdd}},
  Vol.~\bibinfo{volume}{96}. \bibinfo{pages}{226--231}.
\newblock


\bibitem[\protect\citeauthoryear{Faber, K.~Moghaddam, and Wattenhofer}{Faber
  et~al\mbox{.}}{2021}]%
        {faber2021comparing}
\bibfield{author}{\bibinfo{person}{Lukas Faber}, \bibinfo{person}{Amin
  K.~Moghaddam}, {and} \bibinfo{person}{Roger Wattenhofer}.}
  \bibinfo{year}{2021}\natexlab{}.
\newblock \showarticletitle{When Comparing to Ground Truth is Wrong: On
  Evaluating GNN Explanation Methods}. In \bibinfo{booktitle}{\emph{Proceedings
  of the 27th ACM SIGKDD Conference on Knowledge Discovery \& Data Mining}}.
  \bibinfo{pages}{332--341}.
\newblock


\bibitem[\protect\citeauthoryear{Fan, Ma, Li, He, Zhao, Tang, and Yin}{Fan
  et~al\mbox{.}}{2019}]%
        {Fan2019GraphNN}
\bibfield{author}{\bibinfo{person}{Wenqi Fan}, \bibinfo{person}{Y. Ma},
  \bibinfo{person}{Qing Li}, \bibinfo{person}{Yuan He}, \bibinfo{person}{Y.
  Zhao}, \bibinfo{person}{Jiliang Tang}, {and} \bibinfo{person}{D. Yin}.}
  \bibinfo{year}{2019}\natexlab{}.
\newblock \showarticletitle{Graph Neural Networks for Social Recommendation}.
\newblock \bibinfo{journal}{\emph{The World Wide Web Conference}}
  (\bibinfo{year}{2019}).
\newblock


\bibitem[\protect\citeauthoryear{Gilmer, Schoenholz, Riley, Vinyals, and
  Dahl}{Gilmer et~al\mbox{.}}{2017}]%
        {gilmer2017neural}
\bibfield{author}{\bibinfo{person}{Justin Gilmer}, \bibinfo{person}{Samuel~S
  Schoenholz}, \bibinfo{person}{Patrick~F Riley}, \bibinfo{person}{Oriol
  Vinyals}, {and} \bibinfo{person}{George~E Dahl}.}
  \bibinfo{year}{2017}\natexlab{}.
\newblock \showarticletitle{Neural Message Passing for Quantum Chemistry}. In
  \bibinfo{booktitle}{\emph{ICML}}.
\newblock


\bibitem[\protect\citeauthoryear{Hamilton, Ying, and Leskovec}{Hamilton
  et~al\mbox{.}}{2017}]%
        {hamilton2017inductive}
\bibfield{author}{\bibinfo{person}{Will Hamilton}, \bibinfo{person}{Zhitao
  Ying}, {and} \bibinfo{person}{Jure Leskovec}.}
  \bibinfo{year}{2017}\natexlab{}.
\newblock \showarticletitle{Inductive representation learning on large graphs}.
\newblock \bibinfo{journal}{\emph{Advances in neural information processing
  systems}}  \bibinfo{volume}{30} (\bibinfo{year}{2017}).
\newblock


\bibitem[\protect\citeauthoryear{Huang, Yamada, Tian, Singh, Yin, and
  Chang}{Huang et~al\mbox{.}}{2020}]%
        {huang2020graphlime}
\bibfield{author}{\bibinfo{person}{Qiang Huang}, \bibinfo{person}{Makoto
  Yamada}, \bibinfo{person}{Yuan Tian}, \bibinfo{person}{Dinesh Singh},
  \bibinfo{person}{Dawei Yin}, {and} \bibinfo{person}{Yi Chang}.}
  \bibinfo{year}{2020}\natexlab{}.
\newblock \showarticletitle{Graphlime: Local interpretable model explanations
  for graph neural networks}.
\newblock \bibinfo{journal}{\emph{arXiv preprint arXiv:2001.06216}}
  (\bibinfo{year}{2020}).
\newblock


\bibitem[\protect\citeauthoryear{Kipf and Welling}{Kipf and Welling}{2016}]%
        {kipf2016semi}
\bibfield{author}{\bibinfo{person}{Thomas~N Kipf} {and} \bibinfo{person}{Max
  Welling}.} \bibinfo{year}{2016}\natexlab{}.
\newblock \showarticletitle{Semi-supervised classification with graph
  convolutional networks}.
\newblock \bibinfo{journal}{\emph{arXiv preprint arXiv:1609.02907}}
  (\bibinfo{year}{2016}).
\newblock


\bibitem[\protect\citeauthoryear{Lin, Liu, Zhou, Hu, Wang, Zhao, Zheng, Lin,
  Xing, and Liang}{Lin et~al\mbox{.}}{2022}]%
        {lin2022prototypical}
\bibfield{author}{\bibinfo{person}{Shuai Lin}, \bibinfo{person}{Chen Liu},
  \bibinfo{person}{Pan Zhou}, \bibinfo{person}{Zi-Yuan Hu},
  \bibinfo{person}{Shuojia Wang}, \bibinfo{person}{Ruihui Zhao},
  \bibinfo{person}{Yefeng Zheng}, \bibinfo{person}{Liang Lin},
  \bibinfo{person}{Eric Xing}, {and} \bibinfo{person}{Xiaodan Liang}.}
  \bibinfo{year}{2022}\natexlab{}.
\newblock \showarticletitle{Prototypical graph contrastive learning}.
\newblock \bibinfo{journal}{\emph{IEEE Transactions on Neural Networks and
  Learning Systems}} (\bibinfo{year}{2022}).
\newblock


\bibitem[\protect\citeauthoryear{Lin, Lan, and Li}{Lin et~al\mbox{.}}{2021}]%
        {lin2021generative}
\bibfield{author}{\bibinfo{person}{Wanyu Lin}, \bibinfo{person}{Hao Lan}, {and}
  \bibinfo{person}{Baochun Li}.} \bibinfo{year}{2021}\natexlab{}.
\newblock \showarticletitle{Generative causal explanations for graph neural
  networks}. In \bibinfo{booktitle}{\emph{International Conference on Machine
  Learning}}. PMLR, \bibinfo{pages}{6666--6679}.
\newblock


\bibitem[\protect\citeauthoryear{Luo, Cheng, Xu, Yu, Zong, Chen, and Zhang}{Luo
  et~al\mbox{.}}{2020}]%
        {luo2020parameterized}
\bibfield{author}{\bibinfo{person}{Dongsheng Luo}, \bibinfo{person}{Wei Cheng},
  \bibinfo{person}{Dongkuan Xu}, \bibinfo{person}{Wenchao Yu},
  \bibinfo{person}{Bo Zong}, \bibinfo{person}{Haifeng Chen}, {and}
  \bibinfo{person}{Xiang Zhang}.} \bibinfo{year}{2020}\natexlab{}.
\newblock \showarticletitle{Parameterized explainer for graph neural network}.
\newblock \bibinfo{journal}{\emph{Advances in neural information processing
  systems}}  \bibinfo{volume}{33} (\bibinfo{year}{2020}),
  \bibinfo{pages}{19620--19631}.
\newblock


\bibitem[\protect\citeauthoryear{Mansimov, Mahmood, Kang, and Cho}{Mansimov
  et~al\mbox{.}}{2019}]%
        {Mansimov2019MolecularGP}
\bibfield{author}{\bibinfo{person}{Elman Mansimov}, \bibinfo{person}{O.
  Mahmood}, \bibinfo{person}{Seokho Kang}, {and} \bibinfo{person}{Kyunghyun
  Cho}.} \bibinfo{year}{2019}\natexlab{}.
\newblock \showarticletitle{Molecular Geometry Prediction using a Deep
  Generative Graph Neural Network}.
\newblock \bibinfo{journal}{\emph{Scientific Reports}}  \bibinfo{volume}{9}
  (\bibinfo{year}{2019}).
\newblock


\bibitem[\protect\citeauthoryear{Nauta, Trienes, Pathak, Nguyen, Peters,
  Schmitt, Schl{\"o}tterer, van Keulen, and Seifert}{Nauta
  et~al\mbox{.}}{2022}]%
        {nauta2022anecdotal}
\bibfield{author}{\bibinfo{person}{Meike Nauta}, \bibinfo{person}{Jan Trienes},
  \bibinfo{person}{Shreyasi Pathak}, \bibinfo{person}{Elisa Nguyen},
  \bibinfo{person}{Michelle Peters}, \bibinfo{person}{Yasmin Schmitt},
  \bibinfo{person}{J{\"o}rg Schl{\"o}tterer}, \bibinfo{person}{Maurice van
  Keulen}, {and} \bibinfo{person}{Christin Seifert}.}
  \bibinfo{year}{2022}\natexlab{}.
\newblock \showarticletitle{From Anecdotal Evidence to Quantitative Evaluation
  Methods: A Systematic Review on Evaluating Explainable AI}.
\newblock \bibinfo{journal}{\emph{arXiv preprint arXiv:2201.08164}}
  (\bibinfo{year}{2022}).
\newblock


\bibitem[\protect\citeauthoryear{Pope, Kolouri, Rostami, Martin, and
  Hoffmann}{Pope et~al\mbox{.}}{2019}]%
        {pope2019explainability}
\bibfield{author}{\bibinfo{person}{Phillip~E Pope}, \bibinfo{person}{Soheil
  Kolouri}, \bibinfo{person}{Mohammad Rostami}, \bibinfo{person}{Charles~E
  Martin}, {and} \bibinfo{person}{Heiko Hoffmann}.}
  \bibinfo{year}{2019}\natexlab{}.
\newblock \showarticletitle{Explainability methods for graph convolutional
  neural networks}. In \bibinfo{booktitle}{\emph{Proceedings of the IEEE/CVF
  Conference on Computer Vision and Pattern Recognition}}.
  \bibinfo{pages}{10772--10781}.
\newblock


\bibitem[\protect\citeauthoryear{Rao, Zheng, and Yang}{Rao
  et~al\mbox{.}}{2021}]%
        {rao2021quantitative}
\bibfield{author}{\bibinfo{person}{Jiahua Rao}, \bibinfo{person}{Shuangjia
  Zheng}, {and} \bibinfo{person}{Yuedong Yang}.}
  \bibinfo{year}{2021}\natexlab{}.
\newblock \showarticletitle{Quantitative Evaluation of Explainable Graph Neural
  Networks for Molecular Property Prediction}.
\newblock \bibinfo{journal}{\emph{arXiv preprint arXiv:2107.04119}}
  (\bibinfo{year}{2021}).
\newblock


\bibitem[\protect\citeauthoryear{Schnake, Eberle, Lederer, Nakajima,
  Sch{\"u}tt, M{\"u}ller, and Montavon}{Schnake et~al\mbox{.}}{2020}]%
        {schnake2020higher}
\bibfield{author}{\bibinfo{person}{Thomas Schnake}, \bibinfo{person}{Oliver
  Eberle}, \bibinfo{person}{Jonas Lederer}, \bibinfo{person}{Shinichi
  Nakajima}, \bibinfo{person}{Kristof~T Sch{\"u}tt},
  \bibinfo{person}{Klaus-Robert M{\"u}ller}, {and}
  \bibinfo{person}{Gr{\'e}goire Montavon}.} \bibinfo{year}{2020}\natexlab{}.
\newblock \showarticletitle{Higher-order explanations of graph neural networks
  via relevant walks}.
\newblock \bibinfo{journal}{\emph{arXiv preprint arXiv:2006.03589}}
  (\bibinfo{year}{2020}).
\newblock


\bibitem[\protect\citeauthoryear{Shan, Shen, Zhang, Li, and Li}{Shan
  et~al\mbox{.}}{2021}]%
        {shan2021reinforcement}
\bibfield{author}{\bibinfo{person}{Caihua Shan}, \bibinfo{person}{Yifei Shen},
  \bibinfo{person}{Yao Zhang}, \bibinfo{person}{Xiang Li}, {and}
  \bibinfo{person}{Dongsheng Li}.} \bibinfo{year}{2021}\natexlab{}.
\newblock \showarticletitle{Reinforcement Learning Enhanced Explainer for Graph
  Neural Networks}.
\newblock \bibinfo{journal}{\emph{Advances in Neural Information Processing
  Systems}}  \bibinfo{volume}{34} (\bibinfo{year}{2021}).
\newblock


\bibitem[\protect\citeauthoryear{Sorokin and Gurevych}{Sorokin and
  Gurevych}{2018}]%
        {Sorokin2018ModelingSW}
\bibfield{author}{\bibinfo{person}{Daniil Sorokin} {and} \bibinfo{person}{Iryna
  Gurevych}.} \bibinfo{year}{2018}\natexlab{}.
\newblock \showarticletitle{Modeling Semantics with Gated Graph Neural Networks
  for Knowledge Base Question Answering}.
\newblock \bibinfo{journal}{\emph{ArXiv}}  \bibinfo{volume}{abs/1808.04126}
  (\bibinfo{year}{2018}).
\newblock


\bibitem[\protect\citeauthoryear{Tang, Li, and Yu}{Tang et~al\mbox{.}}{2019}]%
        {Tang2019ChebNetEA}
\bibfield{author}{\bibinfo{person}{S. Tang}, \bibinfo{person}{Bo Li}, {and}
  \bibinfo{person}{Haijun Yu}.} \bibinfo{year}{2019}\natexlab{}.
\newblock \showarticletitle{ChebNet: Efficient and Stable Constructions of Deep
  Neural Networks with Rectified Power Units using Chebyshev Approximations}.
\newblock \bibinfo{journal}{\emph{ArXiv}}  \bibinfo{volume}{abs/1911.05467}
  (\bibinfo{year}{2019}).
\newblock


\bibitem[\protect\citeauthoryear{Tsamardinos, Brown, and Aliferis}{Tsamardinos
  et~al\mbox{.}}{2006}]%
        {tsamardinos2006max}
\bibfield{author}{\bibinfo{person}{Ioannis Tsamardinos},
  \bibinfo{person}{Laura~E Brown}, {and} \bibinfo{person}{Constantin~F
  Aliferis}.} \bibinfo{year}{2006}\natexlab{}.
\newblock \showarticletitle{The max-min hill-climbing Bayesian network
  structure learning algorithm}.
\newblock \bibinfo{journal}{\emph{Machine learning}} \bibinfo{volume}{65},
  \bibinfo{number}{1} (\bibinfo{year}{2006}), \bibinfo{pages}{31--78}.
\newblock


\bibitem[\protect\citeauthoryear{Veli{\v{c}}kovi{\'c}, Cucurull, Casanova,
  Romero, Lio, and Bengio}{Veli{\v{c}}kovi{\'c} et~al\mbox{.}}{2017}]%
        {velivckovic2017graph}
\bibfield{author}{\bibinfo{person}{Petar Veli{\v{c}}kovi{\'c}},
  \bibinfo{person}{Guillem Cucurull}, \bibinfo{person}{Arantxa Casanova},
  \bibinfo{person}{Adriana Romero}, \bibinfo{person}{Pietro Lio}, {and}
  \bibinfo{person}{Yoshua Bengio}.} \bibinfo{year}{2017}\natexlab{}.
\newblock \showarticletitle{Graph attention networks}.
\newblock \bibinfo{journal}{\emph{arXiv preprint arXiv:1710.10903}}
  (\bibinfo{year}{2017}).
\newblock


\bibitem[\protect\citeauthoryear{Vu and Thai}{Vu and Thai}{2020}]%
        {vu2020pgm}
\bibfield{author}{\bibinfo{person}{Minh~N Vu} {and} \bibinfo{person}{My~T
  Thai}.} \bibinfo{year}{2020}\natexlab{}.
\newblock \showarticletitle{Pgm-explainer: Probabilistic graphical model
  explanations for graph neural networks}.
\newblock \bibinfo{journal}{\emph{arXiv preprint arXiv:2010.05788}}
  (\bibinfo{year}{2020}).
\newblock


\bibitem[\protect\citeauthoryear{Wang, Jin, Zhang, He, Xu, and Chua}{Wang
  et~al\mbox{.}}{2020a}]%
        {wang2020disentangled}
\bibfield{author}{\bibinfo{person}{Xiang Wang}, \bibinfo{person}{Hongye Jin},
  \bibinfo{person}{An Zhang}, \bibinfo{person}{Xiangnan He},
  \bibinfo{person}{Tong Xu}, {and} \bibinfo{person}{Tat-Seng Chua}.}
  \bibinfo{year}{2020}\natexlab{a}.
\newblock \showarticletitle{Disentangled graph collaborative filtering}. In
  \bibinfo{booktitle}{\emph{Proceedings of the 43rd international ACM SIGIR
  conference on research and development in information retrieval}}.
  \bibinfo{pages}{1001--1010}.
\newblock


\bibitem[\protect\citeauthoryear{Wang, Wu, Zhang, He, and Chua}{Wang
  et~al\mbox{.}}{2020b}]%
        {wang2020causal}
\bibfield{author}{\bibinfo{person}{Xiang Wang}, \bibinfo{person}{Yingxin Wu},
  \bibinfo{person}{An Zhang}, \bibinfo{person}{Xiangnan He}, {and}
  \bibinfo{person}{Tat-seng Chua}.} \bibinfo{year}{2020}\natexlab{b}.
\newblock \showarticletitle{Causal Screening to Interpret Graph Neural
  Networks}.
\newblock  (\bibinfo{year}{2020}).
\newblock


\bibitem[\protect\citeauthoryear{Wu, Wang, Zhang, He, and Chua}{Wu
  et~al\mbox{.}}{2022}]%
        {wu2022discovering}
\bibfield{author}{\bibinfo{person}{Ying-Xin Wu}, \bibinfo{person}{Xiang Wang},
  \bibinfo{person}{An Zhang}, \bibinfo{person}{Xiangnan He}, {and}
  \bibinfo{person}{Tat-Seng Chua}.} \bibinfo{year}{2022}\natexlab{}.
\newblock \showarticletitle{Discovering Invariant Rationales for Graph Neural
  Networks}.
\newblock \bibinfo{journal}{\emph{arXiv preprint arXiv:2201.12872}}
  (\bibinfo{year}{2022}).
\newblock


\bibitem[\protect\citeauthoryear{Xiao, Chen, Guo, Zhuang, and Wang}{Xiao
  et~al\mbox{.}}{2022}]%
        {xiao2022decoupled}
\bibfield{author}{\bibinfo{person}{Teng Xiao}, \bibinfo{person}{Zhengyu Chen},
  \bibinfo{person}{Zhimeng Guo}, \bibinfo{person}{Zeyang Zhuang}, {and}
  \bibinfo{person}{Suhang Wang}.} \bibinfo{year}{2022}\natexlab{}.
\newblock \showarticletitle{Decoupled Self-supervised Learning for
  Non-Homophilous Graphs}.
\newblock \bibinfo{journal}{\emph{arXiv e-prints}} (\bibinfo{year}{2022}),
  \bibinfo{pages}{arXiv--2206}.
\newblock


\bibitem[\protect\citeauthoryear{Xiao, Chen, Wang, and Wang}{Xiao
  et~al\mbox{.}}{2021}]%
        {xiao2021learning}
\bibfield{author}{\bibinfo{person}{Teng Xiao}, \bibinfo{person}{Zhengyu Chen},
  \bibinfo{person}{Donglin Wang}, {and} \bibinfo{person}{Suhang Wang}.}
  \bibinfo{year}{2021}\natexlab{}.
\newblock \showarticletitle{Learning how to propagate messages in graph neural
  networks}. In \bibinfo{booktitle}{\emph{Proceedings of the 27th ACM SIGKDD
  Conference on Knowledge Discovery \& Data Mining}}.
  \bibinfo{pages}{1894--1903}.
\newblock


\bibitem[\protect\citeauthoryear{Xu, Dai, Zhang, and Wang}{Xu
  et~al\mbox{.}}{2022}]%
        {xu2022hp}
\bibfield{author}{\bibinfo{person}{Junjie Xu}, \bibinfo{person}{Enyan Dai},
  \bibinfo{person}{Xiang Zhang}, {and} \bibinfo{person}{Suhang Wang}.}
  \bibinfo{year}{2022}\natexlab{}.
\newblock \showarticletitle{HP-GMN: Graph Memory Networks for Heterophilous
  Graphs}.
\newblock \bibinfo{journal}{\emph{arXiv preprint arXiv:2210.08195}}
  (\bibinfo{year}{2022}).
\newblock


\bibitem[\protect\citeauthoryear{Xu, Hu, Leskovec, and Jegelka}{Xu
  et~al\mbox{.}}{2018}]%
        {xu2018powerful}
\bibfield{author}{\bibinfo{person}{Keyulu Xu}, \bibinfo{person}{Weihua Hu},
  \bibinfo{person}{Jure Leskovec}, {and} \bibinfo{person}{Stefanie Jegelka}.}
  \bibinfo{year}{2018}\natexlab{}.
\newblock \showarticletitle{How powerful are graph neural networks?}
\newblock \bibinfo{journal}{\emph{arXiv preprint arXiv:1810.00826}}
  (\bibinfo{year}{2018}).
\newblock


\bibitem[\protect\citeauthoryear{Ying, Bourgeois, You, Zitnik, and
  Leskovec}{Ying et~al\mbox{.}}{2019}]%
        {ying2019gnnexplainer}
\bibfield{author}{\bibinfo{person}{Rex Ying}, \bibinfo{person}{Dylan
  Bourgeois}, \bibinfo{person}{Jiaxuan You}, \bibinfo{person}{Marinka Zitnik},
  {and} \bibinfo{person}{Jure Leskovec}.} \bibinfo{year}{2019}\natexlab{}.
\newblock \showarticletitle{Gnnexplainer: Generating explanations for graph
  neural networks}.
\newblock \bibinfo{journal}{\emph{Advances in neural information processing
  systems}}  \bibinfo{volume}{32} (\bibinfo{year}{2019}),
  \bibinfo{pages}{9240}.
\newblock


\bibitem[\protect\citeauthoryear{Yuan, Yu, Gui, and Ji}{Yuan
  et~al\mbox{.}}{2020}]%
        {yuan2020explainability}
\bibfield{author}{\bibinfo{person}{Hao Yuan}, \bibinfo{person}{Haiyang Yu},
  \bibinfo{person}{Shurui Gui}, {and} \bibinfo{person}{Shuiwang Ji}.}
  \bibinfo{year}{2020}\natexlab{}.
\newblock \showarticletitle{Explainability in graph neural networks: A
  taxonomic survey}.
\newblock \bibinfo{journal}{\emph{arXiv preprint arXiv:2012.15445}}
  (\bibinfo{year}{2020}).
\newblock


\bibitem[\protect\citeauthoryear{Yuan, Yu, Wang, Li, and Ji}{Yuan
  et~al\mbox{.}}{2021}]%
        {yuan2021explainability}
\bibfield{author}{\bibinfo{person}{Hao Yuan}, \bibinfo{person}{Haiyang Yu},
  \bibinfo{person}{Jie Wang}, \bibinfo{person}{Kang Li}, {and}
  \bibinfo{person}{Shuiwang Ji}.} \bibinfo{year}{2021}\natexlab{}.
\newblock \showarticletitle{On explainability of graph neural networks via
  subgraph explorations}. In \bibinfo{booktitle}{\emph{International Conference
  on Machine Learning}}. PMLR, \bibinfo{pages}{12241--12252}.
\newblock


\bibitem[\protect\citeauthoryear{Zhang and Chen}{Zhang and Chen}{2018}]%
        {zhang2018link}
\bibfield{author}{\bibinfo{person}{Muhan Zhang} {and} \bibinfo{person}{Yixin
  Chen}.} \bibinfo{year}{2018}\natexlab{}.
\newblock \showarticletitle{Link prediction based on graph neural networks}.
\newblock \bibinfo{journal}{\emph{Advances in neural information processing
  systems}}  \bibinfo{volume}{31} (\bibinfo{year}{2018}).
\newblock


\bibitem[\protect\citeauthoryear{Zhang, Liu, Wang, Lu, and Lee}{Zhang
  et~al\mbox{.}}{2021}]%
        {zhang2021protgnn}
\bibfield{author}{\bibinfo{person}{Zaixi Zhang}, \bibinfo{person}{Qi Liu},
  \bibinfo{person}{Hao Wang}, \bibinfo{person}{Chengqiang Lu}, {and}
  \bibinfo{person}{Cheekong Lee}.} \bibinfo{year}{2021}\natexlab{}.
\newblock \showarticletitle{ProtGNN: Towards Self-Explaining Graph Neural
  Networks}.
\newblock \bibinfo{journal}{\emph{arXiv preprint arXiv:2112.00911}}
  (\bibinfo{year}{2021}).
\newblock


\bibitem[\protect\citeauthoryear{Zhao, Tang, Zhang, and Wang}{Zhao
  et~al\mbox{.}}{2020}]%
        {zhao2020semi}
\bibfield{author}{\bibinfo{person}{Tianxiang Zhao}, \bibinfo{person}{Xianfeng
  Tang}, \bibinfo{person}{Xiang Zhang}, {and} \bibinfo{person}{Suhang Wang}.}
  \bibinfo{year}{2020}\natexlab{}.
\newblock \showarticletitle{Semi-Supervised Graph-to-Graph Translation}. In
  \bibinfo{booktitle}{\emph{Proceedings of the 29th ACM International
  Conference on Information \& Knowledge Management}}.
  \bibinfo{pages}{1863--1872}.
\newblock


\bibitem[\protect\citeauthoryear{Zhao, Zhang, and Wang}{Zhao
  et~al\mbox{.}}{2021}]%
        {zhao2021graphsmote}
\bibfield{author}{\bibinfo{person}{Tianxiang Zhao}, \bibinfo{person}{Xiang
  Zhang}, {and} \bibinfo{person}{Suhang Wang}.}
  \bibinfo{year}{2021}\natexlab{}.
\newblock \showarticletitle{GraphSMOTE: Imbalanced Node Classification on
  Graphs with Graph Neural Networks}. In \bibinfo{booktitle}{\emph{Proceedings
  of the Fourteenth ACM International Conference on Web Search and Data
  Mining}}.
\newblock


\bibitem[\protect\citeauthoryear{Zhao, Zhang, and Wang}{Zhao
  et~al\mbox{.}}{2022}]%
        {zhao2022exploring}
\bibfield{author}{\bibinfo{person}{Tianxiang Zhao}, \bibinfo{person}{Xiang
  Zhang}, {and} \bibinfo{person}{Suhang Wang}.}
  \bibinfo{year}{2022}\natexlab{}.
\newblock \showarticletitle{Exploring edge disentanglement for node
  classification}. In \bibinfo{booktitle}{\emph{Proceedings of the ACM Web
  Conference 2022}}. \bibinfo{pages}{1028--1036}.
\newblock


\end{thebibliography}

\newpage 
\appendix

\end{document}